\documentclass[11pt]{article}

\makeatletter
\def\input@path{{latex/}}
\makeatother

\usepackage[final]{acl}

\usepackage{times}
\usepackage{latexsym}

\usepackage[T1]{fontenc}

\usepackage[utf8]{inputenc}

\usepackage{microtype}

\usepackage{inconsolata}

\usepackage{graphicx}
\graphicspath{{figures/}{latex/figures/}}
\usepackage{array}
\usepackage{booktabs}
\usepackage{float}
\usepackage{subcaption}
\usepackage{amsmath}
\usepackage{amssymb}
\usepackage{pifont}
\usepackage{cleveref}
\usepackage{multirow}

\makeatletter
\AddToHook{cmd/appendix/before}{\def\cref@section@alias{appendix}\def\cref@subsection@alias{appendix}}
\makeatother
\crefname{appendix}{Appendix}{Appendices}
\Crefname{appendix}{Appendix}{Appendices}
\crefname{section}{\S}{\S\S}
\Crefname{section}{\S}{\S\S}
\crefname{subsection}{\S}{\S\S}
\Crefname{subsection}{\S}{\S\S}
\crefname{subsubsection}{\S}{\S\S}
\Crefname{subsubsection}{\S}{\S\S}
\AtBeginDocument{%
  \crefformat{section}{\S#2#1#3}\Crefformat{section}{\S#2#1#3}%
  \crefformat{subsection}{\S#2#1#3}\Crefformat{subsection}{\S#2#1#3}%
  \crefformat{subsubsection}{\S#2#1#3}\Crefformat{subsubsection}{\S#2#1#3}%
}

\usepackage[dvipsnames]{xcolor}
\input{macros}
\usepackage[textsize=footnotesize]{todonotes}

\makeatletter
\newcommand*{\iftodonotes}{\if@todonotes@disabled\expandafter\@secondoftwo\else\expandafter\@firstoftwo\fi}
\makeatother

\definecolor{darkblue}{rgb}{0, 0, 0.5}
\hypersetup{
  colorlinks=true,
  citecolor=darkblue,
  linkcolor=darkblue,
  urlcolor=darkblue
}

\newcommand{\cmark}{\ding{51}}
\newcommand{\xmark}{\ding{55}}

\newcommand{\word}[1]{\emph{#1}}

\title{Grammar Control for Language Learning \\
with Sequential Monte Carlo and Small Language Models}

\author{\mdseries %
  \textbf{Kate McCurdy}\textsuperscript{1}
  \hspace{1.5em}
  \textbf{Jacob Hoover Vigly}\textsuperscript{2}
  \hspace{1.5em}
  \textbf{Thomas Hikaru Clark}\textsuperscript{3}
  \\[0.3ex]
  \textbf{Karolina Sta\'nczak}\textsuperscript{4}
  \hspace{1.5em}
  \textbf{Yahya Emara}\textsuperscript{5}
  \hspace{1.5em}
  \textbf{Joshua B. Tenenbaum}\textsuperscript{2,3}
  \\[0.3ex]
  \textbf{Alexander K. Lew}\textsuperscript{6}
  \hspace{1.5em}
  \textbf{Ryan Cotterell}\textsuperscript{4}
  \hspace{1.5em}
  \textbf{Timothy J. O'Donnell}\textsuperscript{2,7,8,9}
  \\[1.2ex]
  \textsuperscript{1}University of G\"{o}ttingen
  \hspace{0.6em}
  \textsuperscript{2}CHI-FRO
  \hspace{0.6em}
  \textsuperscript{3}MIT
  \hspace{0.6em}
  \textsuperscript{4}ETH Z\"{u}rich
  \hspace{0.6em}
  \textsuperscript{5}Cornell University
  \\[0.3ex]
  \textsuperscript{6}Yale University
  \hspace{0.6em}
  \textsuperscript{7}McGill University
  \hspace{0.6em}
  \textsuperscript{8}Mila
  \hspace{0.6em}
  \textsuperscript{9}Canada CIFAR AI Chair
}

\begin{document}
  \maketitle
  \begin{abstract}
    Second language learners benefit from exposure to level-appropriate grammar in
    interactive language use. While prompted frontier language models can usually produce target grammar structures, less resource-intensive models that can be run on device often fail in this task. We use sequential Monte
    Carlo (SMC) to steer dialogue turn generation in small language models
    toward target grammar constructions from an expert-developed
    pedagogical resource. With a minimal set of hand-coded potential functions,
    SMC achieves near-perfect coverage of target grammar, enabling small models to outperform state-of-the-art results from models 140 times larger. We also use code synthesis
    with frontier models to generate new potentials, showing that
    our method scales to hundreds of target grammar skills with no need for additional training or supervised data. %
  \end{abstract}

  \section{Introduction}

  A key task in second language pedagogy is exposing language learners to level-appropriate
  target grammatical structures in diverse natural contexts.
  For example, a learner who
  speaks English at an advanced beginner level (e.g., level A2 in CEFR, the
  Common European Framework of Reference) should be able to understand and use the verb
  \word{enjoy} followed by a verb in its gerund (\word{-ing}) form (e.g., \word{enjoy swimming},
  \word{enjoy cooking}).
  A capable English-language instructor can use this target construction in
  dialogue (e.g., \emph{I enjoy cycling. What exercises do you enjoy?}), encouraging
  learners to practice and develop this grammatical skill---that is, producing grammatical constructions (e.g., \word{enjoy swimming}) rather than ungrammatical word sequences (e.g., \word{enjoy to swim}).
  We would like language models to provide similar interactive support for learners. Whether a model can do so reliably, however, depends heavily on its scale.
  Frontier language models can usually produce learner-appropriate grammatical
  constructions through prompting, but smaller, less resource-intensive models
  often fail to reliably produce grammar targets \citep{sun-etal-2023-evaluating,ribeiro-flucht_framework_2025}.
  Bridging this gap could make efficient, on-device language learning tools more
  feasible, reducing costs and reliance on cloud-based frontier-model APIs while
  broadening access to learning applications.\looseness=-1

  This problem is naturally formulated as a constrained generation task: Given some
  set of level-appropriate grammar skills, we want to constrain language
  model generations such that they contain the target constructions. Pedagogical
  resources such as the English Grammar Profile \citep{okeeffe_english_2017}
  identify a range of grammatical targets at specific proficiency levels. Previous
  research has investigated grammar control via prompting, fine-tuning, reinforcement
  learning, and classifier-based constrained decoding, but none have yet achieved
  reliable elicitation of target constructions from small language models \citep{okano_generating_2023, glandorf-meurers-2024-towards, glandorf_grammar_2025}.

  We can formulate constrained generation as a probabilistic inference problem.
  A language model defines a prior $\prior(\str)$ over strings $\str\in\strings$, where $\vocab$ is an alphabet of tokens.
   Given a \emph{potential} $\potential(\str)$---a binary function that checks whether a string $\str$ satisfies the constraint---we want to sample from the posterior $\frac{\prior(\str)\potential(\str)}{\sum_{\str'\in\strings}\prior(\str')\potential(\str')}$.
  The denominator term sums over all strings, which is intractable; this
  motivates use of approximate inference techniques such as sequential Monte
  Carlo (SMC). SMC builds an importance-weighted population of candidate strings (\emph{particles})
  whose empirical distribution approximates the target posterior. Recent work has
  used SMC to approximate target distributions for constrained generation from language
  models \citep{lew2023sequentialmontecarlosteering, zhao2024probabilisticinferencelanguagemodels, loula2025syntacticsemanticcontrollarge, lipkin2025fastcontrolledgenerationlanguage, xefteri2025syntacticcontrollanguagemodels,grand2025selfsteering,emara2026fasterllminference,golowich2026rejectresamplerepeatunderstanding,kim2025improvingconstrainedlanguagegeneration}.
  Crucially, SMC operates at inference time on any base language model. In our setting, it enables small,
  efficient models to achieve high levels of target grammar coverage without
  additional training. %

  In this work, we specify potential functions via %
  the \GenLMControl library,\footnote{\href{https://genlm.org/genlm-control/}{genlm.org/genlm-control}}
  using short, interpretable, human-readable programs to steer generation. %
  For example,
  a potential function for the \word{enjoy} + gerund construction might parse the
  candidate sentence and return a high score if and only if the verb \word{enjoy}
  is followed by a verb gerund, excluding superficially similar sequences (e.g., \word{enjoy
  something}). Because potential functions can be implemented as explicit
  programs, they can be flexibly developed, modified, and composed. Programmers
  can directly write functions to specify target grammar constructions, which we
  do in Experiment 1. Such programs can also be synthesized by large language models,
  and refined through iterative generation and testing, a technique known as self-steering
  SMC \citep{grand2025selfsteering}; we demonstrate that this approach can scale
  to cover tens (Experiment 2) or hundreds (Experiment 3) of targets. %
  In all cases, potential functions can be inspected and refined by domain
  experts (e.g., instructors or curriculum designers), allowing flexible
  adaptation for specific pedagogical goals. %

  In this work, we apply SMC\footnote{Code: \href{https://github.com/kmccurdy/SMC-L2-grammar}{github.com/kmccurdy/SMC-L2-grammar}} to elicit dialogue turns with target grammar
  constructions %
  from language models with fewer than 10 billion parameters (Figure
  \ref{fig:overview}). %
  We make the following contributions:
  \begin{itemize}
    \item We show that %
    SMC steering can raise the target grammar accuracy of a model with 0.5B parameters from 58\% up to 98\%---outperforming previous state-of-the-art (SotA) from a model 140 times larger---
  while optionally composing with other
      pedagogically-motivated constraints (Experiment 1).%

    \item We show that frontier models can effectively synthesize grammar potential functions through self-steering, and combining up to 6 generated potentials at inference time outperforms previous SotA results (Experiment 2). Moreover, this approach scales to hundreds of targets with no need for supervised data or additional training (Experiment 3). %
  \end{itemize}

  \section{Background}

  Previous research on dialogue adaptation for second language learning has investigated learner-appropriate constraints on vocabulary \citep{tyen-etal-2022-towards,qian_user_2023} and certain grammar categories \citep{okano_generating_2023}. More recently, \citet{ribeiro-flucht_framework_2025} and \citet{glandorf_grammar_2025} explore controlled dialogue generation with a wider range of target grammar constructions grounded in the pedagogical framework of the English Grammar Profile (EGP), and establish SotA performance for specific learner-appropriate grammar targets. As our experiments replicate their setup,
  we first introduce the EGP, %
  and then review these two recent studies in greater depth.

  \subsection{Targets: English Grammar Profile}
  \label{sec:egp}

  \begin{table}[t]
    \centering
    \scriptsize
    \setlength{\tabcolsep}{2pt}
    \begin{tabular}{@{} p{0.8cm} p{3cm} p{3.5cm} %
    @{}}
      \toprule ID (CEFR)                                                                                                                & Can-do statement,  \newline{} describing grammar skill                                                                            & Examples of correct target use \newline{}  in English learner corpora                                                                           \\
      \midrule 258 (A1)                                                                                                                 & FORM: PHRASE + PHRASE \newline Can use single word conjunctions \texttt{(`and, but, or')} to combine phrases.                                                  & My home is very light and very comfortable.                                                                                                %
      \newline I live with my mother and her boyfriend.                                                                                  %
      \newline My house is small but very nice.                                                                                          %
      \\
      \midrule 262 (A1)                                                                                                                 & USE: \texttt{`BUT'}, CONCESSIVE \newline Can use \texttt{`but'} to add an unexpected contrast.                                                                 & I really like my home, it is very small but it is very easy to clean.                                                                      %
      \newline I like my home, it's small but so comfortable. %
      \\
      \midrule 265 (B1)                                                                                                                 & FORM/USE: \texttt{`PLUS'} WITH CLAUSES \newline Can use \texttt{`plus'} to connect clauses and sentences, often to point out a positive addition or advantage. & I would be more than pleased to receive you at home, it's not too big, but we could have a lot of fun, plus you won't have to pay a thing! %
      \newline                                                                                                                           %
      It will be fun because everyone is going to be there. Plus, the park is the best place for a picnic, we can run and just have fun! %
      \\
      \bottomrule
    \end{tabular}
    \caption{Example EGP target grammar skills.}
    \label{tab:ex_egp}
  \end{table}

  We use the English Grammar Profile \citep[EGP;][]{okeeffe_english_2017}, a pedagogical
  resource that lists 1,222 grammar skills for learners of English at specific proficiency
  levels. This resource was developed through analysis of learner corpora, and identifies
  each grammar skill with the proficiency level in the Common European Framework of Reference (CEFR) at which it is typically mastered. A skill is considered
  mastered when learners correctly produce the target grammar construction in
  appropriate contexts.

  \begin{table*}[!ht]
    \centering
    \small
    \begin{tabular}{@{}>{\raggedright\arraybackslash}p{2.2cm}p{0.72\textwidth}@{}}
      \toprule \textbf{Prompt style}      & \textbf{Prompt} \\
      \midrule \textbf{B} (Basic)         & You are a friendly English tutor. I want to practice by 'Discussing enjoyable hobbies'.  \\
      \addlinespace \textbf{T} (Template) & I am a beginner English student. I want to practice patterns with to and -ing by Discussing hobbies with `enjoy + ing'. I want to ensure that I can use `enjoy' followed by the `-ing' form. For example, `I enjoy eating, I enjoy doing, I enjoy going…'. Let's do a Q\&A exercise. You (the AI) are a friendly English tutor. Tell me What you enjoy doing. Then ask me the same with short questions. (`I enjoy reading. What do you enjoy doing?', `When I go out I enjoy dancing. How about you?') \\
      \bottomrule
    \end{tabular}
    \caption{Partial basic vs.\ template prompts for EGP skill 1082 (\emph{enjoy
    + -ing}) with the topic ``hobbies'' \citep{ribeiro-flucht_framework_2025}. For
    complete prompts and further details, see \cref{app:expt1_constraints}. }
    \label{tab:enjoy-prompts}
  \end{table*}

  EGP grammar skills are expressed as ``can do'' statements with supporting
  examples, and may target either a specific grammatical \emph{form}, its \emph{use}
  in a particular discourse context, or both.\footnote{For more details, see the EGP website: \href{https://englishprofile.org}{englishprofile.org}}
  Table \ref{tab:ex_egp} gives some example skills from the category ``coordinating
  conjunctions.'' Note that, while target grammar constructions are expressed clearly,
  some skills are open to ambiguity in interpretation; for instance, evaluators
  of EGP ID 262 may differ in judgments of an ``unexpected contrast.'' While we aim
  for consistent evaluation, such ambiguities lead to unavoidable annotator
  variation and noise.

  \subsection{Prior work}
  \label{sec:prior}

  EGP skills have been used to define target grammar constructions in two recent studies establishing SotA for pedagogical grammar control, which our experiments build upon.
  \citet{ribeiro-flucht_framework_2025} assess tutoring dialogue scenarios in which
  the language model is prompted to use a target grammar construction while engaging an L2 English student in dialogue on a specific topic. They identify 5 grammar skills at proficiency level A2, and test each skill with different topics. The authors
  compare two prompting approaches: a \textbf{basic (B)} prompt with relatively minimal instructions, and a more elaborate \textbf{template}-based \textbf{(T)} prompt with detailed instructions and few-shot examples (Table \ref{tab:enjoy-prompts}). They also run a pedagogically-informed grammar check, and prompt models to regenerate any response with grammar beyond a B1 proficiency level.
  They find that template prompting substantially increases accurate production of
  target grammar constructions across a range of model scales, with a peak
  performance of 91\% from Llama 3.3 70B, rising to 96.3\% with an additional validation step.\looseness=-1

  \citet{glandorf_grammar_2025} assess naturalistic dialogue generation in Llama
  3.1 8B. They use a wider range of 29 EGP grammar skills, spanning proficiency
  levels from A1 to C1, and additionally test combinations of up to 6 grammar targets
  applied to a single generation. Instead of different prompts, the authors compare
  two technical approaches to improving model performance: supervised \textbf{fine-tuning},
  and constrained generation using \textbf{guided decoding} with \emph{Future Discriminators for Generation} \citep[FUDGE;][]{yang-klein-2021-fudge}.
  They find that a \textbf{hybrid} combining both approaches achieves the
  highest performance, with an average accuracy of 64\% across grammar target
  combinations, ranging from 87\% with one target to 50\% with 6. %

  \section{Methods}

  \begin{figure*}[!ht]
    \centering
    \includegraphics[width=\linewidth, trim={20pt 75pt 20pt 75pt}, clip]{
      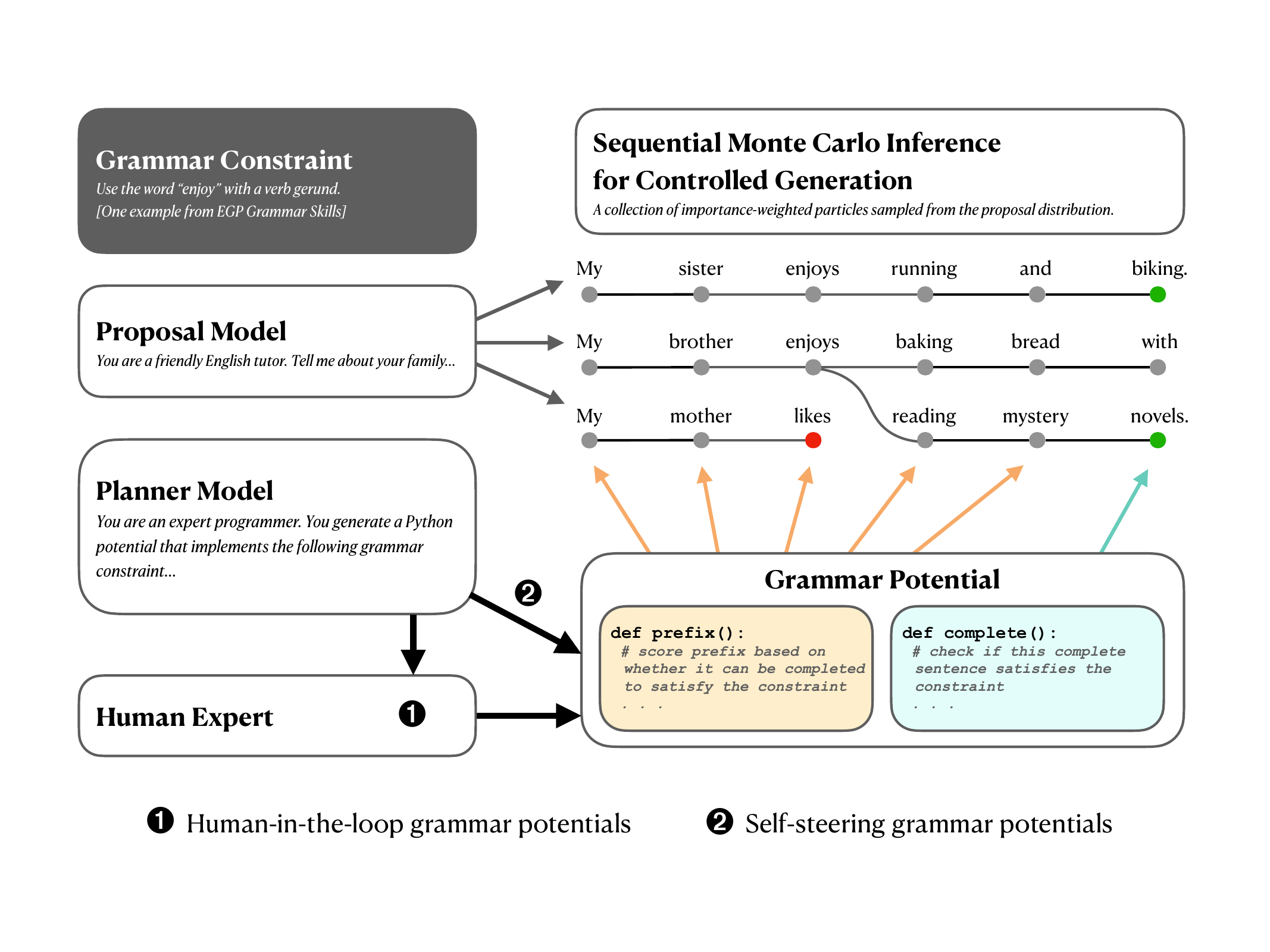
    }
    \caption{Approach Overview. We use explicit grammar constraints, encoded in
    potential functions, to steer generation in a small language model with sequential
    Monte Carlo. }
    \label{fig:overview}
  \end{figure*}

  In this section, we explain how we implement potential functions to steer generation
  toward target grammar constructions from EGP skills, and how those generations
  are evaluated. Experiments 1 and 2 focus on the specific grammar skills
  targeted in prior work, while Experiment 3 covers the full set of skills in the
  English Grammar Profile.

  \subsection{Generation}
  \label{sec:generation}

  \paragraph{Problem formulation.}
  Let $\vocab$ be an \emph{alphabet}, i.e., a finite, nonempty set of tokens, and let $\prior$ be a language model, i.e., a probability distribution over $\strings$, the set of all finite strings over $\vocab$. Given a binary
  potential $\potential\colon\strings\to\{0,1\}$,
  define the target posterior $\target(\str)\defeq
  \prior(\str)\potential(\str)/Z$, where the normalizing constant (marginal
  likelihood) $Z\defeq\sum_{\stralt\in\strings}\prior(\stralt)\potential(\stralt)$
  sums over all strings $\stralt\in\strings$. This normalizing constant cannot generally be tractably
  computed, so sampling from the target distribution requires approximate inference
  techniques. We approximate $\target$ with sequential Monte Carlo (SMC), which
  maintains a collection of $\numParticles$ \emph{particles} iteratively extended
  from a \emph{proposal} language model $\proposal$.\footnote{The choice of proposal is a degree of freedom for the practitioner. In all experiments here, we choose to use the prior, $\prior$.}
  During generation, particles are iteratively extended, reweighted against the target,
  and occasionally resampled to focus computation on promising partial strings (see
  \Cref{app:smc_detail} for details).
  SMC yields a posterior approximation as a weighted set of particles; in this
  work we always select the top-weighted particle for evaluation.

  \paragraph{Implementation of grammar potentials.}
  We use \GenLMControl (v0.3.1) to implement SMC for constrained generation.
  This library allows us to express potentials as Python programs (Figure
  \ref{fig:overview}). In the \GenLMControl API, a potential exposes two methods:
  (1) a \texttt{complete} method, called once to evaluate
  $\potential$ on a \emph{finished} string, and (2) a \texttt{prefix} method, called at every generation step to score the current \emph{partial} sequence. Two considerations
  motivate the separation. First, a \texttt{complete} method may require a
  completed phrase to correctly confirm the presence of a particular grammatical
  construct. Second, \texttt{complete} may be expensive to compute; evaluating
  it only once per particle, at termination, keeps decoding fast. We exploit the
  design freedom of \texttt{prefix} %
  to provide a cheap incremental steering signal that rewards partial sequences with candidate
  cues (e.g., a penalty that reduces %
  when a regular expression detects the string \emph{enjoy}), while
  \texttt{complete} verifies the full target structure on the finished sequence (e.g., parsing the sentence with \emph{enjoy} to confirm a dependent verb gerund). We use the spaCy library \citep[v3.8.14;][]{spacy} to implement strict grammar checking in potentials' \texttt{complete} method; for examples, see Table \ref{tab:expt1_constraints} (\Cref{app:expt1_constraints}).

  \paragraph{Implementation of vocabulary potential.} Motivated by the level-appropriate constraints explored in prior work, we additionally implement a GenLM potential to restrict generation to a defined vocabulary list. This potential runs an identical binary check for both \texttt{prefix} and \texttt{complete}, penalizing any particle with words outside the vocabulary whitelist. As GenLM supports flexibly composing multiple potentials, we combine this vocabulary restriction with grammar targets in Experiment 1.

  \paragraph{Self-steering potential function generation.}
  In Experiments 2 and 3, we assess automated generation of potential functions
  using the \emph{self-steering language models} approach proposed by \citet{grand2025selfsteering}.
  In this technique, a large frontier language model (the Planner model; cf. Figure
  \ref{fig:overview}) is prompted to %
  write a program implementing the target constraint, and then immediately
  tests the program by using it to steer generation in a smaller language model.
  If the test generation does not satisfy the constraint---i.e., in our case,
  does not produce the target grammar construction specified by a given EGP
  skill---then the Planner model can iterate by generating and testing a new
  potential function. Our goal in self-steering
  is to evaluate the feasibility of scaling generation of potential functions.
  Hand-coding each potential may take too much effort to be practical in large-scale pedagogical applications, but automated generation with self-steering can reduce this burden; a single run of the Planner model produces interpretable Python programs that are easy to correct if needed, and these programs along with a small language model are the only requirements to run at inference time.%

  \subsection{Evaluation}
  \label{sec:eval}

  \paragraph{Grammar coverage.}
  We use an LLM (GPT-5-nano) to evaluate grammar target presence for generations
  in all experiments. Specifically, we pass in the full EGP specification for a particular target grammar construct (e.g.,
  Table \ref{tab:ex_egp}) with a set of texts and prompt the model to assign True
  or False to each text based on the presence of the target construct. As this approach differs from the evaluation methods of the prior work we reproduce in
  Experiments 1 and 2, %
  we validate our pipeline by running on their original generated texts.
  Table \ref{tab:eval_validation} in \cref{app:expt1_constraints} compares
  our grammar annotation approach to previously reported results on the same set
  of generations. While we find minor discrepancies between our evaluation and original reported numbers, overall trends (i.e., relative performance of models) appear consistent.
  For additional validation, we use an alternative LLM (Claude Haiku 4.5) as a
  judge for Experiment 1, and find 94.5\% agreement with GPT-5-nano; see \Cref{app:annotation-comparison}
  for details and example disagreements.

  \paragraph{Speed, memory, and quality.}
  For other metrics, we reproduce the evaluation of \citet{glandorf_grammar_2025}.
  We assess the same dialogue generation quality metrics (Appropriateness,
  Relevance, Content Richness, Grammatical Correctness) using LLM-as-judge, with
  the same model (GPT-4o) and prompts as their experiment. \citet{ribeiro-flucht_framework_2025}
  use human annotators to evaluate generation appropriateness; we instead use LLMs
  as described here. We also follow \citet{glandorf_grammar_2025} in calculating
  generation speed in words per minute, as a more intuitively relevant measure
  than token-based generation speed. Second-language learners have an estimated reading
  speed of 70 to 90 words per minute \citep{iwahori2008developing}, so we take this range as a lower bound for acceptable generation speed. We additionally assess the computational demands of SMC for both GPU- and CPU-based settings with memory profiling (\Cref{app:memory_profile}).

  \paragraph{Human validation.} As human judgment is particularly important to second-language-learning applications, we additionally validate both our grammar and quality evaluation pipelines through human comparison. We take a randomly sampled subset of 99 SMC generations from Experiments 1 and 2, and recruit online crowdworkers to judge target grammar presence (11 raters) and generation quality (13 raters). Results are broadly consistent with our automated pipeline; see details in \Cref{app:annotation-comparison}.

  \paragraph{Baseline.}
  SMC constrains generation by utilizing inference-time compute to generate and resample particles in parallel.
  For a fair comparison with comparable inference budgets, we compare SMC generation in Experiments 1
  and 2 to two baselines: 1) standard language model decoding with best-of-$\numParticles$ selection and 2) rejection sampling (sampling complete generations, and rejecting those that lack the target construction). For each SMC run with $\numParticles$
  particles, we sample $\numParticles$ generations from the same language model with the
  same prompt; we implement the rejection sampling baseline by removing incremental steering (i.e., setting the potential's \texttt{.prefix()} method to accept all prefixes) and retaining only the final grammar check (the \texttt{.complete()} method) to reject generations that do not satisfy the hard grammar constraint. For both SMC generation and rejection sampling, we select the top-weighted particle; for our standard language model decoding baseline, we take the top generated sequence measured by
  lowest average per-token surprisal. See detailed rejection sampling results in \Cref{app:expt1_constraints}.

  \section{Experiments}

  \subsection{Experiment 1}\label{sec:experiment1}

  This experiment replicates the grammar control task posed by \citet{ribeiro-flucht_framework_2025},
  using SMC to steer generation with hand-coded symbolic potential functions for
  5 target grammar constructions. We explore three research questions.
  The first compares directly to \citeauthor{ribeiro-flucht_framework_2025}'s results:
  beyond the original prompting effect, can SMC steering boost accurate
  production of target grammar constructions in Llama 3.1 8B?
  The second measures performance across model scales: can SMC additionally boost target grammar production in even smaller models?
  Our third research question explores how SMC potentials can flexibly compose multiple relevant constraints. This is motivated by our pedagogical application: in order to support gradual learning, L2 language instructors often deliberately introduce new grammatical constructions using only vocabulary words familiar to their students.
  We therefore ask whether SMC-based steering can maintain target accuracy and response quality while using a restricted vocabulary, i.e., only the set of words typically known to A2-level English learners.\looseness=-1

  \begin{figure*}[t]
    \centering
    \includegraphics[width=0.95\linewidth]{
      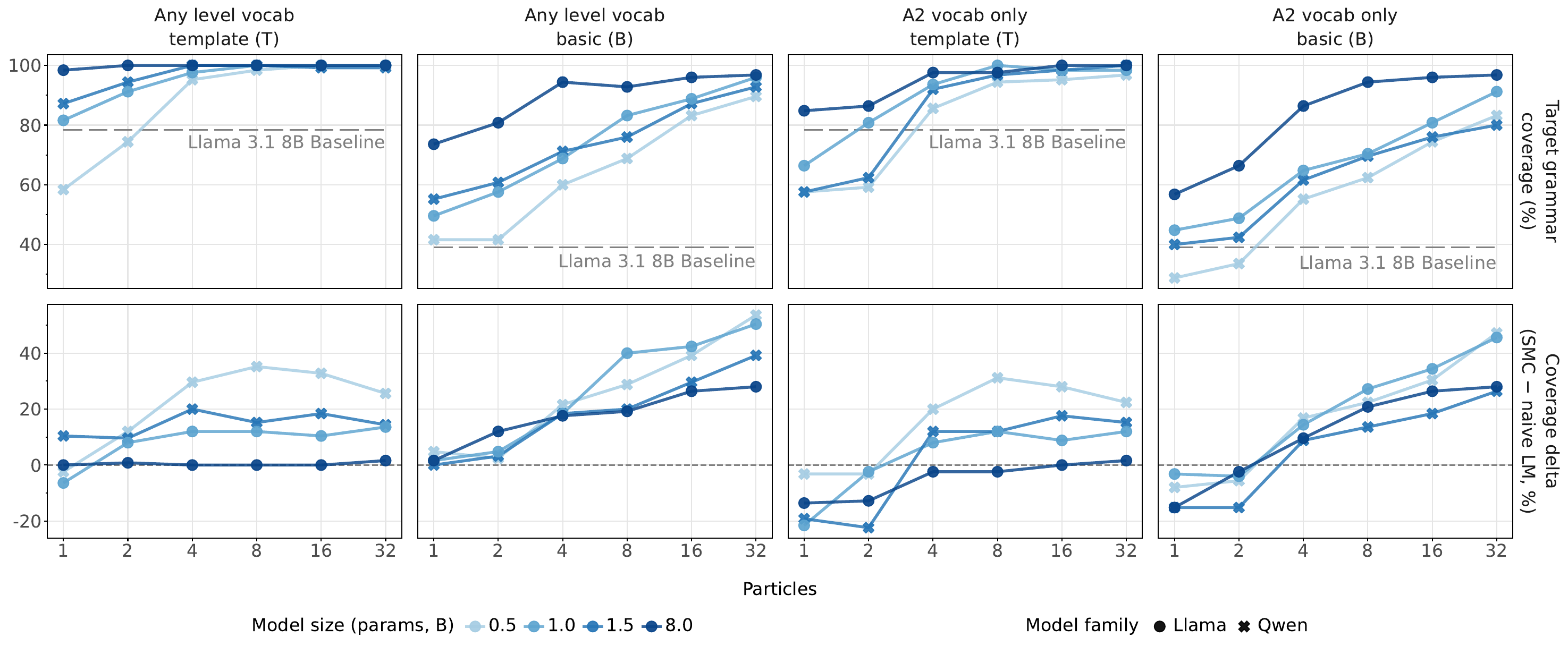
    }
    \caption{Experiment 1: Target grammar coverage by number of particles and model
    scale. Marker shapes denote model family (Llama vs. Qwen). Grammar coverage
    improves both with increasing particle count and increasing model size (top row),
    with smaller models showing the largest improvement relative to language model baseline (bottom
    row). Top row dotted line shows Llama 3.1 8B accuracy reported from the original study.
    }
    \label{fig:expt1_scale}
  \end{figure*}

  \subsubsection{Task}

  \citet{ribeiro-flucht_framework_2025} identify 5 target grammar skills, and for each skill, 5 different
  communicative scenarios where that grammar construction could naturally appear
  in conversation.
  Each context-skill combination is assessed with both the Basic (B) and Template (T) prompt in 5 distinct runs, totaling 2 $\times$ 125 generations per model.
  We focus comparison on their smallest tested model, Llama 3.1 8B, which rises from 39\% accuracy with Basic to 78\% with Template prompting.
  As discussed in \cref{sec:prior}, we use an LLM-based
  evaluation pipeline validated on the original study's generated texts (\Cref{app:expt1_constraints}).
  In terms of accuracy, our replication appears slightly lower than the original paper; however,
  this may reflect noise as discussed in \cref{sec:egp}, so we compare to their original reported accuracy rates.\looseness=-1

  \subsubsection{Our approach}

  \paragraph{Language models.} %
  We assess generation with four small open-source instruction-tuned models. From the Llama family \citep{grattafiori2024llama3herdmodels}, we evaluate Llama 3.1 8B and Llama 3.2 1B; from Qwen \citep{yang2024qwen2technicalreport}, we evaluate Qwen 2 1.5B and 0.5B.

  \paragraph{Grammar potentials.}
  We write potential functions for each of the 5 grammar skills as described in
  \cref{sec:generation}. Table \ref{tab:expt1_constraints} in \Cref{app:expt1_constraints}
  lists the five EGP grammar skills targeted in Experiment 1, and illustrates
  how we implement potential functions to constrain generation.

  \paragraph{Vocabulary potential.}
  We explore a further pedagogical capacity of SMC: restricting generation to a
  constrained vocabulary appropriate to the learner's proficiency level, as described in \cref{sec:generation}. Using a
  related pedagogical resource from the Cambridge English Profile initiative \citep{english_vocabulary_profile},
  we construct a vocabulary list comprising only words identified with an A2 proficiency
  level, yielding approximately 3,400 unique word forms. %

  \subsubsection{Results}

  \paragraph{Target grammar accuracy.}
  We find that SMC combined with detailed template (T) prompting enables small
  models to achieve near-perfect target grammar coverage (Figure \ref{fig:expt1_scale}).
  This improvement is largest for smaller models: 
  while Llama 3.1 8B reaches
  ceiling accuracy with template prompting alone, increasing the particle budget
  from 1 to 8 pushes accuracy from 58\% to 98\% in the smallest model, Qwen 2 0.5B.\footnote{We see similar trends when comparing to the rejection sampling baseline (Figure \ref{fig:expt1_rejection} and Table \ref{tab:rejection-sampling} in \Cref{app:expt1_constraints}).} Notably, this is an improvement on the best overall performance
  reported by \citet{ribeiro-flucht_framework_2025}, i.e., 96\% accuracy for Llama
  3.3 70B; with inference-time SMC-steered generation,
  this SotA result is exceeded by a model 140 times smaller.

  \paragraph{Vocabulary restriction and quality.}
  We see that increasing particle budget enables similarly high performance even
  when the vocabulary is constrained to a fixed list of A2-level words. For
  example, Qwen 2 0.5B achieves 95\% accuracy with 4 particles and no vocabulary
  restriction, but this drops to 85\% when the vocabulary restriction is added; however,
  raising the particle budget to 8 brings accuracy to 94\% even with restricted vocabulary.
  Higher particle budgets also improve response quality across the board, but the
  effect is more pronounced for A2-vocabulary-only generations (Figure
  \ref{fig:expt1_quality}, \cref{app:expt1_constraints}). In terms of speed,
  increasing particle budgets slows down generation, but rates remain acceptable
  (i.e., greater than 100 words per minute) %
  up to 32 particles---with the one exception of Llama 3.1 8B, which generates 60 words per minute at 32 particles (Figure \ref{fig:expt1_speed},
  \cref{app:expt1_constraints}). These results demonstrate the key trade-off: we
  can compensate for smaller model size and capability, including additional
  pedagogically-motivated restrictions, with increased computation at inference time via SMC.

  \begin{table*}[!ht] \small
    \centering
    \begin{tabular}{cc|rrrrr|r| rrrr}
      \toprule
      \multirow{6}{*}{\rotatebox{90}{\textbf{\citeauthor{glandorf_grammar_2025}}}} & & \multicolumn{5}{l}{\textbf{Number of Constraints}} &             & \multicolumn{4}{l}{\textbf{Average Quality Rating}} \\
      & & 1                                                  & 2           & 3                                                  & 4           & 6           & \textbf{Mean} & App. & C.R. & G.C. & Rel. \\
      & Llama 3.1 8B               & \small 70.6                                        & \small 62.0 & \small 48.0                                        & \small 46.5 & \small 42.0 & 53.8         & \textbf{4.45} & \textbf{4.35} & \textbf{2.90} & \textbf{4.92} \\
      & + fine-tuning              & \small 76.0                                        & \small 61.5 & \small 49.3                                        & \small 49.2 & \small 44.7 & 56.2         & 4.09 & 3.98 & 2.29 & 4.70 \\
      & + decoding                 & \small 81.6                                        & \small \textbf{72.2} & \small 65.7                                        & \small \textbf{60.8} & \small \textbf{65.3} & \textbf{69.1}          & 4.31 & 4.27 & 2.80 & 4.87  \\
      & + both (hybrid)                    & \small \textbf{86.5}                                        & \small 69.8 & \small \textbf{66.7}                                        & \small 60.0 & \small 59.5 & 68.5          & 3.45 & 3.32 & 2.02 & 4.13 \\
      \midrule
     \multirow{8}{*}{\rotatebox{90}{\textbf{SMC (Ours)}}}  & \textbf{Particles} \\
      \addlinespace
      & 1            & 63.1                                               & 52.0        & 48.0                                               & 44.8        & 38.2        & 49.2          & \textbf{4.12} & \textbf{3.11} & \textbf{4.56} & \textbf{4.20} \\
      & 2                          & 73.4                                               & 63.0        & 47.0                                               & 45.3        & 38.0        & 53.3          & 4.10 & 3.10 & 4.55 & 4.10 \\
      & 4                          & 79.8                                               & 74.0        & 60.7                                               & 62.8        & 60.3        & 67.5          & 3.96 & 3.06 & 4.45 & 4.03 \\
      & 8                          & 82.5                                               & 77.8        & 64.3                                               & 67.0        & 66.7        & 71.6          & 3.94 & 3.05 & 4.46 & 4.04 \\
      & 16                         & 84.5                                               & 85.3        & 64.6                                               & 70.0        & 65.7        & 74.0          & 3.98 & 3.06 & 4.48 & 4.02 \\
      & 32                         & \textbf{86.6}                                               & \textbf{86.3}        & \textbf{74.3}                                               & \textbf{75.0}        & \textbf{68.2}        & \textbf{78.1}        & 3.94 & 3.03 & 4.50 & 4.02 \\
       \bottomrule
    \end{tabular}
    \caption{Experiment 2: Llama 3.1 8B (middle) accuracy (\% target coverage) across
    grammar skills; (right) average quality metrics (Appropriateness, Content
    Richness, Grammatical Correctness, Relevance). Top block shows results from
    \citet[][Task 1]{glandorf_grammar_2025}, using higher accuracy numbers from our
    evaluation pipeline (\cref{app:expt1_constraints}); bottom shows SMC generation at different particle budgets. %
    With 8 particles, SMC outperforms previous approaches in producing up to 6 target grammar constructions.
    }
    \label{tab:expt2}
  \end{table*}

  \subsection{Experiment 2}

  Experiment 2 applies our methods to the grammar control task posed by \citet[][Task 1]{glandorf_grammar_2025}. %
  In this case, instead of hand-coding potential functions as in Experiment 1, we use self-steering
  with a Planner language model \citep{grand2025selfsteering} to generate and
  iteratively revise potentials for the 29 targeted grammar skills. We investigate %
  whether SMC with generated potentials improves target accuracy
  relative to previous approaches. %

  \subsubsection{Task}

  \citet{glandorf_grammar_2025} assess controlled generation within naturalistic
  dialogues, with four conversational turns as context preceding the target generation.
  For each generation, the authors randomly sample context from a natural dialogue dataset,
  and then randomly sample 1, 2, 3, 4, or 6 EGP grammar skills to target, out of
  the set of 29. The language model is then prompted with the preceding dialogue
  context and target grammar constructions. For an example prompt and more
  detailed experiment description, see \Cref{app:expt2_details}.

  \citet{glandorf_grammar_2025} explore three different strategies to increase
  grammar control in Llama 3.1 8B. First, they train independent BERT-based
  classifiers for each grammar skill, retaining those with at least 80\%
  accuracy on a set of hand-annotated grammar judgments. These classifiers are then
  used for two distinct purposes. First, the classifiers are used to extract dialogue
  turns which contain target grammar constructions, creating a dataset for
  supervised \textbf{fine-tuning}. Second, the classifiers are additionally trained
  as ``future discriminators'' on the fine-tuned data: given a generated prefix sequence,
  they estimate the probability that the complete sentence will contain the
  target construction. These discriminators are then used to steer \textbf{guided
  decoding} with FUDGE \citep{yang-klein-2021-fudge}. %
  Finally, the \textbf{hybrid} approach combines both, using future
  discriminators to guide decoding on the fine-tuned Llama model. %

  \citeauthor{glandorf_grammar_2025}'s FUDGE-based decoding is similar in spirit to our SMC approach: both reweight a pretrained LM at each generation step using a future-looking signal that approximates the partial sequence's expected future constraint-satisfaction. However, the two methods differ in some important respects. For one, their future discriminators are neural classifiers trained per grammar skill on labeled examples, whereas our prefix potentials are short interpretable Python programs requiring no per-skill data (see \cref{app:smc_detail}). Moreover, in \citeauthor{glandorf_grammar_2025}'s setup FUDGE produces a single sampled trajectory, via a locally guided decoding strategy with top-$k$ and epsilon pruning; we instead sample from the LM and reweight by the potential across $\numParticles$ particles with adaptive resampling. This lets us trade inference compute for accuracy by scaling the number of particles, as shown in Experiment 1.

  \subsubsection{Our approach}

  \paragraph{Self-steering potential generation.} %
  We use self-steered code synthesis following \citet{grand2025selfsteering}, %
  with GPT-5.4 as our Planner model. We prompt the Planner with detailed instructions
  using Experiment 1's hand-coded potentials as few-shot examples. Once the
  potential is generated, it is used to steer generation in Llama 3.1 8B with 16
  particles. The Planner then assesses whether generation contains the target; if
  not, the potential is regenerated up to 5 iterations. %

  \paragraph{Programmatic and manual repair.}
  Despite the various checks built into our self-steering pipeline, we find that
  the Planner model can still generate invalid or low-quality grammar potential functions;
  however, these are easily corrected retroactively. %
  We approach potential repair in two ways. First, we run a script over all potentials to automatically fix common
  code errors in handling spaCy attributes. Second, we identify 8 specific grammar
  skills with consistently low accuracy in initial runs, and manually revise the
  potentials to better fit their targets. %

  \subsubsection{Results}

  For ease of comparison, rather than sampling anew, we replicate the exact dialogue constraint
  combinations and contexts used by \citet{glandorf_grammar_2025} in
  Task 1. In contrast to the previous experiment, our pipeline here consistently
  assigns \emph{higher} target accuracy to the same text generations (\Cref{app:expt2_details}). %
  In the same spirit of fair comparison, we recognize possible evaluation noise
  and report the higher accuracy numbers from our pipeline (Table \ref{tab:expt2}).%

  At 4 particles, Llama 3.1 8B with SMC outperforms \citeauthor{glandorf_grammar_2025}'s fine-tuned model; %
  at 8 particles, it bests guided decoding, the other inference-time approach; and
  at 32 particles, it outperforms the hybrid model combining both. %
  The advantage is especially noticeable for cases where generation must satisfy
  multiple grammar constraints.
  These findings point toward the flexibility and scalability of SMC: steered by auto-generated and lightly edited potential functions, we find that Llama 3.1 8B can reliably produce up to 6 target grammar constructions out of the box, outperforming the SotA results of \citeauthor{glandorf_grammar_2025}'s hybrid model with no additional training on the base model.

       \begin{table*}[!ht]
    \centering
    \small
    \begin{tabular}{c|lrrrrrr}
      \toprule
      & \textbf{Proficiency Level (CEFR)} & \textbf{A1} & \textbf{A2} & \textbf{B1} & \textbf{B2} & \textbf{C1} & \textbf{C2} \\
      \midrule                                    %
      & Number of grammar skills                      & 109         & 288         & 338         & 242         & 129         & 112         \\
      \addlinespace                               %
      GPT-4o-mini & One-shot accuracy       & 84.1        & 76.9        & 77.4        & 69.5        & 77.3        & 73.8        \\
     & Planner for Llama 3.1 8B SMC (16)               & 86.0        & 79.6        & 76.7        & 70.0        & 79.0        & 65.9        \\
      \addlinespace
      GPT-5.4   & One-shot accuracy           & 95.4        & 97.3        & 96.5        & 96.6        & 93.8        & 95.8        \\
      & Planner for Llama 3.1 8B SMC (16)          & 88.7        & 81.2        & 81.6        & 72.2        & 80.1        & 74.6        \\
      \bottomrule
    \end{tabular}
    \caption{Experiment 3: target grammar coverage by proficiency level for 1,218
    EGP grammar constructions. Using grammar potentials generated by a small
    Planner model (GPT-4o-mini), Llama 3.1 8B with SMC (16 particles) achieves comparable performance
    to one-shot generation from the Planner model. Potentials generated by a larger Planner model (GPT-5.4) improve Llama-SMC performance, but not enough to equal the larger Planner
    on its own.
    }
    \label{tab:expt3}
  \end{table*}

  \subsection{Experiment 3}

  \paragraph{Task.}
  As a proof of concept, we apply the self-steering approach to scale potential generation
  to all 1,222 grammar skills in the EGP. We experiment with two different
  Planner models: GPT-4o-mini, a small low-latency frontier model, and GPT-5.4,
  OpenAI's flagship model at the time these experiments were conducted. We use both models
  in two capacities: a) as Planner models generating grammar potentials to steer
  Llama 3.1 8B, and b) as comparative performance baselines with one-shot accuracy. %
  For all potentials, we run the same script to fix common spaCy errors used in Experiment
  2, but perform no further correction.

  \paragraph{Results.}
  Overall, we find that the self-steering approach scales successfully: of the 1,222
  grammar skills in the EGP, Planner model GPT-5.4 generates viable potential functions
  for 99\%, with only 4 constraints excluded for nontrivial errors. We use these
  potentials to steer Llama 3.1 8B using SMC with 16 particles. We find that,
  while performance remains below the skyline of the most capable Planner (GPT-5.4),
  it nevertheless matches the smaller frontier Planner model GPT-4o-mini (Table
  \ref{tab:expt3}). %
  Using high-quality potentials generated from GPT-5.4, SMC elicits robustly high accuracy from Llama 3.1 8B across diverse EGP grammar targets, exceeding 70\% even for high proficiency levels with complex grammar skills. This illustrates the viability and efficiency of scaling SMC: we achieve broad coverage without any supervised data, using only a large Planner model to generate potentials from EGP grammar skills, and a small language model to use those potentials during inference.

  \section{Conclusion}

  We have shown that SMC-based constrained generation can elicit target
  grammatical constructions from small language models with highly accurate
  coverage, outperforming both much larger models that rely on prompting alone,
  and comparable small models with dedicated fine-tuning. A core contribution of
  our framework is the explicit representation of grammatical knowledge in human-readable
  potential functions: these can be hand-coded by domain experts, synthesized by
  a frontier model, or some combination of the two. This flexibility lets our approach
  scale effectively while enabling transparent specification and correction if needed.
  Our approach opens new possibilities for pedagogical applications of small
  language models in low-resource settings, broadening accessibility for the world's language learners.

  \section*{Limitations} Our experiments have several limitations. First, we evaluate generation using %
  single dialogue turns rather than extended multi-turn interactions in a learning context; a full user study with language learners would be needed to confirm the pedagogical value of this approach in practice. Second, while SMC achieves high grammar coverage, increasing the particle budget slows generation, and at 32 particles Llama 3.1 8B falls below the minimum reading speed we consider acceptable. This suggests further work is needed to improve inference efficiency, for example through better proposal distributions or particle pruning. Third, our potential approach depends on mature NLP tooling (spaCy) and an expert-developed pedagogical resource (the EGP), both of which are predominantly available for English; extending the approach to other languages would require developing equivalent resources. Finally, the self-steering pipeline still requires some human intervention, both programmatic and manual repair of generated potentials. To support more efficient scaling, future work should identify ways to improve the potential generation pipeline.

  \paragraph{AI Usage.} We use AI agents for coding support, and lightly in writing and revision. We have thoroughly reviewed all AI suggestions, and take full responsibility for the text. %

  \IfFileExists{latex/acl.sty}
    {\bibliography{latex/colm2026_conference,latex/custom}}
    {\bibliography{colm2026_conference,custom}}

  \clearpage

  \appendix

  \section{SMC for constrained generation}%
  \label{app:smc_detail}

  This appendix provides additional detail on the sequential Monte Carlo
  procedure described in \cref{sec:generation}, and its implementation in the \GenLMControl
  library we use. The background on SMC presented here is standard, applied to our setting of constrained generation from language models following the treatments of \citet{lew2023sequentialmontecarlosteering,loula2025syntacticsemanticcontrollarge}
  and \citet{lipkin2025fastcontrolledgenerationlanguage}, to which we refer the
  reader for more complete coverage.

  \paragraph{Foundations.}
  Let $\vocab$ be an \emph{alphabet}, i.e., a finite, nonempty set of tokens. A \emph{language model} $\lm$
  is a probability distribution over $\strings$, the set of finite strings over
  $\vocab$. Define the \emph{prefix probability} of $\str\in\strings$ under
  $\lm$ as\footnote{We write $\str\prefixOf\str'$ if $\str$ is a prefix of
  $\str'$.}
  \begin{equation}
    \label{eq:prefix_probability}\prelm(\str)\defeq\sum_{\str'\in\strings}\indicator
    {\str\prefixOf\str'}\lm(\str'),
  \end{equation}
  and the \emph{conditional prefix probability} of $\tok'\in\vocab\cup\{\eos\}$
  given $\str\in\strings$ as
  \begin{equation}
    \prelm(\tok'\mid\str) \defeq
    \begin{cases}
      {\lm(\str)/\prelm(\str)}         & \text{if } \tok'=\eos, \\
      {\prelm(\str\tok')/\prelm(\str)} & \text{otherwise}.
    \end{cases}
    \label{eq:conditional_prefix_probability}
  \end{equation}
  This lets us write the following autoregressive factorization of $\lm(\str)$
  for any $\str\in\strings$.
  \begin{equation}
    \lm(\str) = \prelm(\eos\mid\str) \prod_{t=1}^{|\str|}\prelm(\tok_{t}\mid\str_{<t}
    ). \label{eq:autoregressive}
  \end{equation}

  \paragraph{Controlled generation setting.}
  In our setting, for a \emph{prior} language model $\prior$ and a \emph{(complete)
  potential} $\potential\colon\strings\to\Rpos$,\footnote{Note, in our application we
  focus on the special case of constrained generation, where potentials are
  binary functions $\potential\colon\strings\to\{0,1\}$.} we define the target (posterior)
  distribution as
  \begin{equation}
  \target(\str) \defeq \frac{\prior(\str)\potential(\str)}{Z},
  \end{equation}
  with
  $Z \defeq \sum_{\str'\in\strings}\prior (\str')\potential(\str')$ being the normalizing constant.
  SMC approximates the intractable $\target$ by sampling from a tractable
  proposal distribution $\proposal$, a language model over the same alphabet,\footnote{We
  will also require that $\proposal(\str)=0\implies\prior(\str)=0$ holds for all
  $\str\in\strings$ (so that the proposal is supported anywhere the target is supported, which is required for the IS weights of \cref{eq:full-weight} to be well-defined).} and denote its prefix probability $\preProposal(\str)$
  and conditional prefix probability $\preProposal(\tok'\mid\str)$, defined as in \cref{eq:prefix_probability,eq:conditional_prefix_probability,eq:autoregressive}. In the simplest case $\proposal
  =\prior$, but $\proposal$ may differ from the prior in general.

  \paragraph{Particle approximation.}
  SMC is an instance of \emph{importance sampling} (IS), which produces a
  discrete random estimate of an intractable target distribution by reweighting
  samples (called \emph{particles}) drawn from a tractable proposal. Sample particles $\str\psup{m}\iid\proposal$, for $m=1,\ldots,\numParticles$, and assign each
  particle an \emph{importance weight}
  \begin{equation}
    \Weight\psup{m} \defeq \frac{\prior(\str\psup{m})\,\potential(\str\psup{m})}{\proposal(\str\psup{m})}
    = Z\,\frac{\target(\str\psup{m})}{\proposal(\str\psup{m})},
    \label{eq:full-weight}
  \end{equation}
  proportional to the density ratio $\target/\proposal$ (this weight can be thought of as correcting for the fact
  that the particle was sampled from $\proposal$ rather than from $\target$). The weighted collection of particles $\left(\str\psup{m}, \Weight\psup{m}\right)_{m=1}^{\numParticles}$ allows us to define an unbiased estimate of the normalizing constant $Z$ as
  \begin{equation}
    \estZ \defeq \frac{1}{\numParticles}\sum_{m=1}^{\numParticles}\Weight\psup{m},
  \end{equation}
  and form the consistent (but biased) \emph{self-normalized importance sampling estimate} of the target distribution $\target$, as
  \begin{equation}
    \estTarget(\str) \defeq \frac{\frac{1}{\numParticles}\sum_{m=1}^{\numParticles}\Weight\psup{m}\,\indicator{\str=\str\psup{m}}}{\estZ}.
    \label{eq:particle-approx}
  \end{equation}
  SMC extends IS by constructing the particles \emph{sequentially}:
  Rather than sampling complete strings $\str\psup{m}\sim\proposal$ and then
  computing the full weight ratio of \cref{eq:full-weight}, SMC extends partial
  strings token by token, updating each particle's weight incrementally, as described below.

  \paragraph{Prefix potential.}
  To enable incremental reweighting at each generation step, we additionally
  specify a \emph{prefix potential} $\prePotential$. Given a complete potential
  $\potential$, a function $\prePotential\colon\strings\to\Rpos$ is a prefix potential
  for $\potential$ if it satisfies the following \emph{monotonicity constraint}: $\prePotential
  (\str)=0 \implies \prePotential(\str')=\potential(\str')=0$ for every $\str'\in
  \strings$ with $\str\prefixOf\str'$. Intuitively, that is, if the prefix potential zeroes
  out a partial sequence, then it must remain zero on all completions of that
  partial sequence, and those completions must already have zero weight under the
  complete potential. From $\prePotential$ we define the \emph{incremental
  prefix potential} of $\tok'\in\vocab\cup\{\eos\}$ given $\str\in\strings$,
  analogously to the conditional prefix probability:
  \begin{equation}
    \prePotential(\tok'\mid\str) \defeq
    \begin{cases}
      {\potential(\str)/\prePotential(\str)}         & \text{if }\tok'=\eos, \\[8pt]
      {\prePotential(\str\tok')/\prePotential(\str)} & \text{otherwise}.
    \end{cases}
    \label{eq:incremental-prefix-potential}
  \end{equation}
  \Cref{eq:incremental-prefix-potential} is only needed when $\prePotential(\str)>0$: by the monotonicity constraint, a particle with $\prePotential(\str)=0$ already has weight zero, and every completion of it has zero potential, so it can be discarded.
  Without loss of generality, we take $\prePotential(\epsilon)=1$.\footnote{Given
  any prefix potential $\prePotential$ satisfying the monotonicity constraint,
  the rescaled function $\prePotential'(\str)\defeq\prePotential(\str)/\prePotential(\epsilon)$
  also satisfies it and has $\prePotential'(\epsilon)=1$. The SMC weight updates
  under $\prePotential$ versus $\prePotential'$ produce final particle weights
  that differ only by a global constant, so the self-normalized estimator of
  \cref{eq:particle-approx} is invariant to this choice.} The monotonicity constraint then ensures that
  $\potential$ admits an autoregressive factorization through these incremental
  prefix potentials, analogous to \cref{eq:autoregressive}.

  Within the monotonicity constraint, $\prePotential$ is a design choice:
  Choosing $\prePotential$ to track $\potential$'s structure on partial
  sequences (e.g., rewarding partial sentences containing the target construction's
  lexical cues) reduces the variance of the SMC weights without changing the
  target. The variance-minimizing choice is the \emph{optimal twist function}
  $\prePotential^{\star}(\str) = \mathbb{E}_{\str'\sim\prior}[\potential(\str')\mid
  \str\prefixOf\str']$, but since this is generally intractable,
  $\prePotential$ may be designed by hand or learned \citep[see][]{zhao2024probabilisticinferencelanguagemodels,loula2025syntacticsemanticcontrollarge}.

  \paragraph{SMC: Extend and Resample.}
  SMC produces the particle approximation of \cref{eq:particle-approx} by
  iterating two steps until all particles terminate.

  \textbf{(1) Extend.} Each particle samples a next token (or EOS)
  $\tok_{t}\psup{m}\sim \preProposal(\cdot\mid \str_{<t}\psup{m})$. Its
  \emph{importance weight at step $t$}, $\Weight_t\psup{m}$, is then defined
  via the following recurrence relation, with $\Weight_{0}\psup{m}\defeq 1$:
  \begin{equation}
    \Weight_{t}\psup{m}\;=\; \Weight_{t-1}\psup{m}\cdot \frac{\prePrior(\tok_{t}\psup{m}\mid \str_{<t}\psup{m})
    \cdot \prePotential(\tok_{t}\psup{m}\mid \str_{<t}\psup{m})}{\preProposal(\tok_{t}\psup{m}\mid
    \str_{<t}\psup{m})}. \label{eq:smc-weight}
  \end{equation}
  The numerator of the correction factor combines the prior LM with the
  incremental prefix potential; the denominator is the proposal.
  Sampling $\tok_{t}\psup{m}=\eos$ terminates
  generation in a particular particle; via \cref{eq:incremental-prefix-potential},
  the step's weight factor specializes to a final reweighting by
  $\potential/\prePotential$ times the LM/proposal importance correction $\prePrior
  /\preProposal$ ($=1$ when $\proposal=\prior$). In the absence of
  resampling (step 2, below), chaining \cref{eq:smc-weight} via the chain rules for $\prior$,
  $\potential$, and $\proposal$ telescopes to the IS weight of
  \cref{eq:full-weight} at termination (SMC then reduces to sequential importance
  sampling). With resampling, individual cumulative weights are reset
  and no longer equal the IS weights, but the empirical estimator of
  \cref{eq:particle-approx} remains a consistent estimate of $\target$.

  \textbf{(2) Resample.} When the normalized effective sample size
  \begin{equation}
    \mathrm{ESS}_{t} \defeq \frac{\big(\sum_{m=1}^{\numParticles}\Weight_{t}\psup{m}\big)^{2}}{\numParticles \sum_{m=1}^{\numParticles}\big(\Weight_{t}\psup{m}\big)^{2}}
    \label{eq:ess}
  \end{equation}
  drops below some \emph{resampling threshold} (a global hyperparameter of the SMC algorithm, set to $0.5$ by default in \GenLMControl), particles are stochastically pruned and replicated in
  proportion to their weights, and all weights are reset to the average
  $\bar{\Weight}_{t}=\tfrac{1}{\numParticles}\sum_{m=1}^{\numParticles}\Weight_{t}\psup{m}$. This
  preserves the unnormalized total weight while concentrating particles on
  promising partial sequences (those scored higher by $\prePotential$).

  \paragraph{\GenLMControl implementation.}
  The framework above maps directly onto the \GenLMControl library's potential API,
  which exposes a potential via two methods, for scoring $\str\in\strings$ (following
  \cref{eq:incremental-prefix-potential}):
  \begin{itemize}
    \item \texttt{prefix}: realizes the prefix potential $\prePotential(\str)$.
      Called \emph{once per particle per step}, to compute the incremental
      prefix potential.
    \item \texttt{complete}: realizes the complete potential $\potential(\str)$.
      Called \emph{only once per particle}, when $\eos$ is sampled.
  \end{itemize}
  This structure is useful when the complete potential is expensive to evaluate
  (e.g., requires a full dependency parse, as in our experiments): The expensive
  \texttt{complete} evaluation is deferred until termination, while a cheaper
  \texttt{prefix} method is evaluated at every step.

  Note, \citet{loula2025syntacticsemanticcontrollarge} draw a further distinction
  between \emph{efficient} potentials, which can be incorporated into the proposal
  at the token level (requiring evaluation on every token in the alphabet in
  order to normalize the proposal's conditional prefix probability distribution),
  and \emph{expensive} potentials, used as \emph{critics} that reweight particles
  after sampling from the proposal, as in \cref{eq:smc-weight}. This choice is reflected
  in the \GenLMControl API with the option to incorporate a potential directly into
  the proposal, or pass it as a critic. In this work we use all potentials as critics.
  We note that \GenLMControl also provides Adaptive Weighted Rejection Sampling
  \citep[AWRS;][]{lipkin2025fastcontrolledgenerationlanguage}, an efficient
  token-level algorithm that draws unbiased samples from the local product of a
  language model and a hard token-level constraint; integrating AWRS into the proposal
  step we leave to future work.%

  \section{Replication Details}
  \label{app:expt1_constraints}

  \subsection{Experiment 1}

  Table \ref{tab:expt1_constraints} presents all five grammar skills targeted in
  Experiment 1, and how we implement them using the \GenLMControl API. Table
  \ref{tab:enjoy-prompts-compare} gives complete prompts used in Experiment 1, and
  compares our prompts and terms to the original study. %

\begin{figure*}
    \centering
    \includegraphics[width=0.75\linewidth]{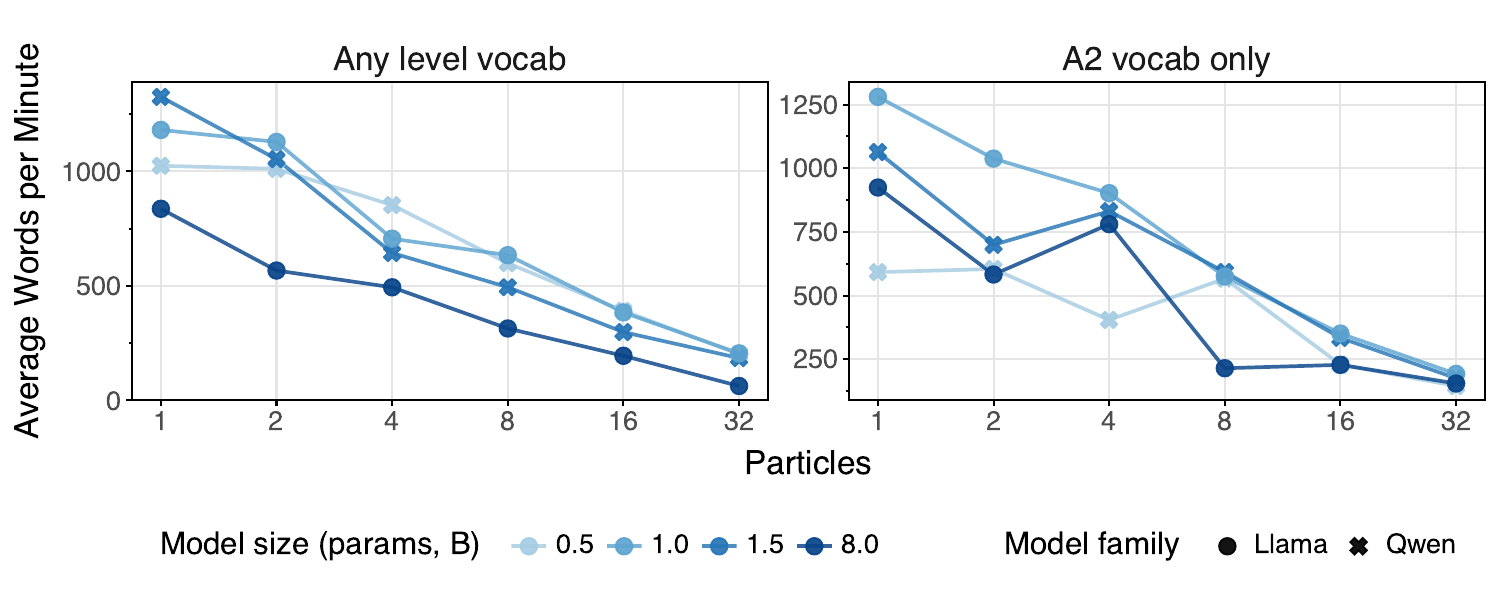}
    \caption{Experiment 1: Average generation speed by vocabulary constraint and
    number of SMC particles.}
    \label{fig:expt1_speed}
  \end{figure*}

\begin{figure*}
    \centering
    \includegraphics[width=0.75\linewidth]{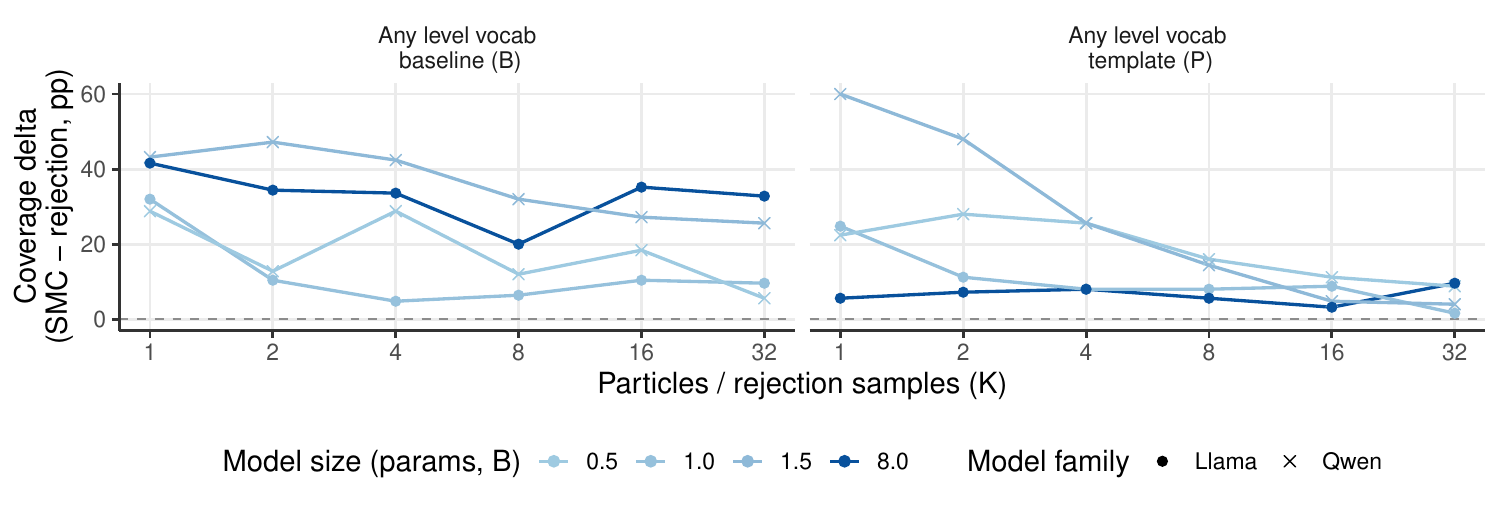}
    \caption{Experiment 1: SMC improvement in target grammar production over rejection sampling baseline.}
    \label{fig:expt1_rejection}
  \end{figure*}

  \citet{ribeiro-flucht_framework_2025} assess target grammar coverage with POLKE
  \citep{sagirov2025polke}, an annotation pipeline customized for the EGP. Our
  preliminary exploration with the open POLKE API found it poorly suited to our
  high-demand needs, e.g., a tendency to crash with repeated calls. Table \ref{tab:eval_validation:e1}
  shows that our LLM-based evaluation of their generated dialogue data consistently
  assesses lower per-turn accuracy, but overall reproduces their key findings: use
  of the template prompt \textbf{T}, especially with their additional verification
  step \textbf{T + V}, substantially boosts target grammar coverage over the baseline
  prompt \textbf{B}, but neither small nor larger language models approach full target
  grammar coverage. Figure \ref{fig:expt1_speed} shows average generation speed
  for Experiment 1. Note that increasing particles slows generation, but speed remains
  within acceptable bounds for all but the largest model; with 32 particles, every model except Llama 3.1 8B still generates more than
  100 words per minute.
  Figure \ref{fig:expt1_rejection} and Table \ref{tab:rejection-sampling} report results from the rejection sampling baseline.

  \subsection{Experiment 2}
  \label{app:expt2_details}

  Task 1 of \citet{glandorf_grammar_2025} focuses on 29 grammar skills drawn from
  3 EGP subcategories (``superlatives'', ``would'', and ``negation'') across all proficiency
  levels. This task assesses controlled generation within naturalistic dialogues,
  with four conversational turns as context preceding the target generation. For
  each task, the authors randomly sample a context from one of two dialogue datasets:
  \textit{Document Grounded Dataset} \citep{zhou-etal-2018-dataset} and \textit{Topical-Chat}
  \citep{gopalakrishnan_topical-chat_2019}. They then randomly sample 1, 2, or 3
  EGP subcategories, and then 1 or 2 grammar skills from each category, so that 1,
  2, 3, 4, or 6 grammatical items are targeted. For each individual generation
  task, the language model receives a template prompt listing the EGP Subcategory,
  Guideword, Can-do statement, and CEFR Level of each targeted grammar skill, and
  is instructed to use all grammatical items in the dialogue's next turn. Table
  \ref{tab:expt2_prompt} provides an example dialogue turn prompt.

  \citet{glandorf_grammar_2025} train independent BERT-based classifiers for
  each grammar skill. These classifiers are used both to evaluate target grammar
  coverage in generation---target grammar structures are evaluated as present when
  detected by a classifier model---and to steer language generation in their constrained
  decoding and hybrid models. Table \ref{tab:eval_validation:e2} reports their
  results for the base model Llama 3.1 8B and three different strategies for
  grammar control: fine-tuning with prompt templates for the target grammar skills,
  classifier-steered constrained decoding with FUDGE \citep{yang-klein-2021-fudge},
  and a hybrid approach with constrained decoding on a fine-tuned model, which achieved
  the best performance overall. The same table also reports results on these
  generations using our LLM-as-judge evaluation pipeline. In contrast to the previous
  experiment, our pipeline here consistently assigns \emph{higher} target accuracy
  to the same text generations, especially the multiple-constraint conditions of
  the constrained decoding model. Nevertheless, we see the same overall
  directional effects as originally reported: highest single-constraint accuracy
  from the hybrid model, and robustly decreasing target coverage with increasing
  number of constraints. Table \ref{tab:rejection-sampling} reports results from the rejection sampling baseline.

  \begin{table*}
    \centering
    \small
    \begin{subtable}{0.45\textwidth}
      \centering
      \begin{tabular}{lrrr}
        \toprule \textbf{Model}      & \textbf{B} & \textbf{T} & \textbf{T + V}          \\
        \midrule Llama 3.1 8B        & 39.0       & 78.4       & \multicolumn{1}{c}{---} \\
        \multicolumn{1}{r}{our eval} & 32.6       & 77.6       &                         \\
        \addlinespace                 %
        Llama 3.3 70B                & 37.0       & 91.4       & 96.3                    \\
        \multicolumn{1}{r}{our eval} & 31.7       & 84.8       & 85.8                    \\
        \addlinespace                 %
        DeepSeek V3                  & 34.2       & 87.8       & 95.0                    \\
        \multicolumn{1}{r}{our eval} & 28.6       & 81.0       & 84.2                    \\
        \addlinespace                 %
        GPT-4o                       & 35.5       & 80.5       & 91.5                    \\
        \multicolumn{1}{r}{our eval} & 25.8       & 75.0       & 81.1                    \\
        \bottomrule
      \end{tabular}
      \subcaption{\citet[][Table 3]{ribeiro-flucht_framework_2025}. \textbf{B} is Basic prompt, \textbf{T} is Template prompt, and \textbf{T + V} is Template with an additional inference-time Validation step. }
      \label{tab:eval_validation:e1}
    \end{subtable}
    \hfill
    \begin{subtable}{0.45\textwidth}
      \centering
      \setlength{\tabcolsep}{3pt}
      \begin{tabular}{lrrrrrr}
        \toprule \textbf{N. constraints} & 1           & 2           & 3           & 4           & 6           & \textbf{Mean} \\
        \midrule Llama 3.1 8B            & \small 60.8 & \small 47.5 & \small 38.0 & \small 37.2 & \small 32.3 & 43.2          \\
        \multicolumn{1}{r}{our eval}     & \small 70.6 & \small 62.0 & \small 48.0 & \small 46.5 & \small 42.0 & 53.8          \\
        \addlinespace                     %
        + fine-tuning                    & \small 69.3 & \small 55.5 & \small 46.3 & \small 45.0 & \small 41.2 & 51.5          \\
        \multicolumn{1}{r}{our eval}     & \small 76.0 & \small 61.5 & \small 49.3 & \small 49.2 & \small 44.7 & 56.2          \\
        \addlinespace                     %
        + decoding                       & \small 77.9 & \small 61.5 & \small 51.3 & \small 51.7 & \small 53.8 & 59.3          \\
        \multicolumn{1}{r}{our eval}     & \small 81.6 & \small 72.2 & \small 65.7 & \small 60.8 & \small 65.3 & 69.1          \\
        \addlinespace                     %
        + both                           & \small 86.7 & \small 66.0 & \small 62.7 & \small 53.2 & \small 49.8 & 63.7          \\
        \multicolumn{1}{r}{our eval}     & \small 86.5 & \small 69.8 & \small 66.7 & \small 60.0 & \small 59.5 & 68.5          \\
        \bottomrule
      \end{tabular}
      \subcaption{\citet[][Table 2]{glandorf_grammar_2025}. Number refers to number of unique constraints applied to generation task.}
      \label{tab:eval_validation:e2}
    \end{subtable}
    \caption{Validating our evaluation pipeline. We use GPT-5-nano for evaluation
    where prior work uses specialized classifiers. Here, we show reported results
    for the constrained generation tasks reproduced in Experiments 1 and 2,
    along with accuracy from our evaluation pipeline on the same set of
    generated texts.}
    \label{tab:eval_validation}
  \end{table*}

  \begin{table*}
    \centering
    \small
    \setlength{\tabcolsep}{4pt}
    \begin{tabular}{@{}c p{2.9cm} p{3.5cm} | p{2.9cm} p{3.5cm}@{}}
      \toprule EGP entry                                                                  &                                                                                                                &                                                                                                                                            & Potential function                                                                              &                                                                                                 \\
      EGP ID                                                                              & Can-do statement                                                                                               & Examples                                                                                                                                   & \texttt{.prefix()}                                                                              & \texttt{.complete()}                                                                            \\
      \midrule 115                                                                        & Can use a range of degree adverbs to modify common gradable adjectives.                                        & I'm so happy because you can come!\newline It's really important for me, because I need to do some homework.                               &                                                                                                  %
      Reduce penalty for degree adverb (really, very, so, too, quite, extremely, pretty). & String contains some adjective (\texttt{tag\_ = ADJ}) with a dependent degree adverb (\texttt{dep\_ = advmod}). \\
      \midrule 786                                                                        & Can use the affirmative form with an increasing range of verbs.                                                & I forgot to tell you some details about tomorrow.                                                                                           %
      \newline I spent about \pounds 35 on all those clothes.                              %
                                                                                          & Reduce penalty for past-verb-like string (\texttt{[a-zA-Z]+ed}, was, were, went, got, ...).                    & String contains token with \texttt{tag\_ = VBD}.                                                                                            \\
      \midrule 874                                                                        & Can use the present simple with adverbs of indefinite frequency.                                               & People usually try to eat their breakfast                                                                                                   %
      before 9 am.\newline They always make special things for their celebrations.        & Reduce penalty for frequency adverb (usually, sometimes, always, often, never, rarely, seldom, frequently).    & String contains some present-tense main verb (\texttt{tag\_ = \{VBP,VBZ,VB\}}) with a dependent frequency adverb (\texttt{dep\_ = advmod}). \\
      \midrule 930                                                                        & Can use \emph{it} as an empty subject                                                                           %
      in subject position, particularly with weather or time.                             & It's warm and windy.\newline Now it's 3 pm.                                                                    & Reduce penalty for \texttt{it} and \texttt{it + (is/was/'s)}.                                                                              & String contains token with \texttt{lemma = be} and dependent \emph{it} (\texttt{dep\_ = nsubj}). \\
      \midrule 1082                                                                       & Can use \emph{enjoy} followed by the \emph{-ing} form.                                                         & We enjoy playing the piano and singing together.\newline You know I really enjoy listening to music.                                       & Reduce penalty for \texttt{enjoy} and \texttt{enjoy + word+ing}.                                & String contains \emph{enjoy} with some verb gerund (\texttt{tag\_ = VBG}) as head or dependent. \\
      \bottomrule
    \end{tabular}
    \caption{Grammar constraints in Experiment 1, all at CEFR level A2. ID, ``can
    do'' statement, and examples are taken from the English Grammar Profile. Rightmost
    columns illustrate how we implement these constraints as potential functions.
    Prefix methods use regular expressions for fast steering, while complete methods
    use dependency parses to enforce correct target grammar.}
    \label{tab:expt1_constraints}
  \end{table*}

  \begin{table*}
    \centering
    \small
    \begin{tabular}{@{}>{\raggedright\arraybackslash}p{2.4cm}>{\raggedright\arraybackslash}p{2.2cm}p{0.58\textwidth}@{}}
      \toprule \textbf{Prompt style} & \textbf{Study}                        & \textbf{Prompt}                                                                                                                                                                                                                                                                                                                                                                                                                                                                                                                                                                                                                                                                                                                                         \\
      \midrule \textbf{B} (Baseline)          & \citet{ribeiro-flucht_framework_2025} & You are a friendly English tutor. I want to practice by 'Discussing enjoyable hobbies'. Please use direct, short and simple sentences and easy words. Output only your next turn. Now let's naturally start our conversation. \par\smallskip User: Hello! System:                                                                                                                                                                                                                                                                                                                                                                                                                                                                                       \\
      \textbf{B} (Basic)                      & Ours                                  & You are a friendly English tutor. I want to practice by `Discussing hobbies with `enjoy + ing''. Please use direct, short and simple sentences and easy words. Output only your next turn. Now let's naturally start our conversation. \par\smallskip User: Hello! Tell me about yourself and the topic, and then ask about me.                                                                                                                                                                                                                                                                                                                                                                                                                         \\
      \addlinespace \textbf{P} (Prompt)       & \citet{ribeiro-flucht_framework_2025} & I am a beginner English student. I want to practice patterns\_with to and -ing by Discussing hobbies with 'enjoy + ing'. I want to ensure that I can use 'enjoy' followed by the '-ing' form. For example, 'I enjoy eating, I enjoy doing, I enjoy going…'. Let's do a Q\&A exercise. You (the AI) are a friendly English tutor. Tell me What you enjoy doing. Then ask me the same with short questions. ('I enjoy reading. What do you enjoy doing?', 'When I go out I enjoy dancing. How about you?') Please use direct, short and simple sentences and easy words. Output only your next turn. Now let's naturally start our conversation. \par\smallskip User: Hello! System:                                                                        \\
      \textbf{T} (Template)                   & Ours                                  & I am a A2 English student. I want to practice enjoy + ing pattern (FORM: `ENJOY' + `-ING') by Discussing hobbies with `enjoy + ing'. I want to ensure that I can use `enjoy' followed by the `-ing' form.. For example, I enjoy eating, I enjoy doing, I enjoy going\ldots{}. Let's do a Q\&A exercise. You (the AI) are a friendly English tutor. Tell me What you enjoy doing. Then ask me the same with short questions. (`I enjoy reading. What do you enjoy doing?', `When I go out I enjoy dancing. How about you?') Please use direct, short and simple sentences and easy words. Output only your next turn. Now let's naturally start our conversation. \par\smallskip User: Hello! Tell me about yourself and the topic, and then ask about me. \\
      \bottomrule
    \end{tabular}
    \caption{Complete basic vs.\ template prompts for EGP skill 1082 (\emph{enjoy
    + -ing}) with the topic ``hobbies.'' As shown here, our Experiment 1 prompts have
    minor changes from the original, and we refer to the prompt categories with
    different terms.}
    \label{tab:enjoy-prompts-compare}
  \end{table*}

  \section{Annotation Validation}
  \label{app:annotation-comparison}

  \subsection{LLM validation}

  To validate our annotations, we reran Experiment~1
  evaluation with a second LLM, Claude Haiku 4.5, on the same set of generations
  originally annotated by GPT-5-nano. Both models received an identical prompt: the
  full EGP specification for the target construct (cf.\ \Cref{tab:ex_egp})
  together with the set of generated texts, with instructions to label each text
  as containing the construct or not. We find that the two models agree 94.5\% of
  the time. As expected, the disagreements concentrate on the smallest models at
  the lowest particle counts, where generations are borderline-grammatical. In \Cref{tab:annotator-disagreements},
  we present representative examples of model disagreements.

  \subsection{Human validation}

  We obtain additional human judgments on a random sample of generations. %
  Data collection was conducted by authors from ETH Z\"urich; in compliance with the institution's ethics commission, only anonymized responses are collected without identifiable personal data.
  We recruit online annotators using the Prolific platform and pay them a rate equivalent to \$12/hour. Participants are asked for consent (Table \ref{tab:human-consent}) and provided study-appropriate instructions (Table \ref{tab:human-instructions}).

  From Experiment 1, we sample 2 (out of 5) topic prompts for each of 3 (out of 5) grammar constraints. We only consider generations from the more successful template (T) prompt condition, but sample from each of the 2 vocabulary conditions (unrestricted vs. A2-only). We compare generations from our largest model (Llama 3.1 8B) and smallest (Qwen 2 0.5B). Given overall high performance in this task, we assess generations from the lower span of particle compute budgets, sampling generations with 1, 4, and 16 particles. This yields 72 individual generations to be rated.

  From Experiment 2, we consider only generations from Llama 3.1 8B, and focus sampling across cases with 1, 3, and 6 target grammar constraints. As this task is harder, we shift to evaluating a range of higher particle counts: 2, 8, and 32. We evaluate 3 individual cases respectively from each of these conditions, yielding 27 individual generations.

\begin{figure}
    \centering
    \includegraphics[width=\linewidth]{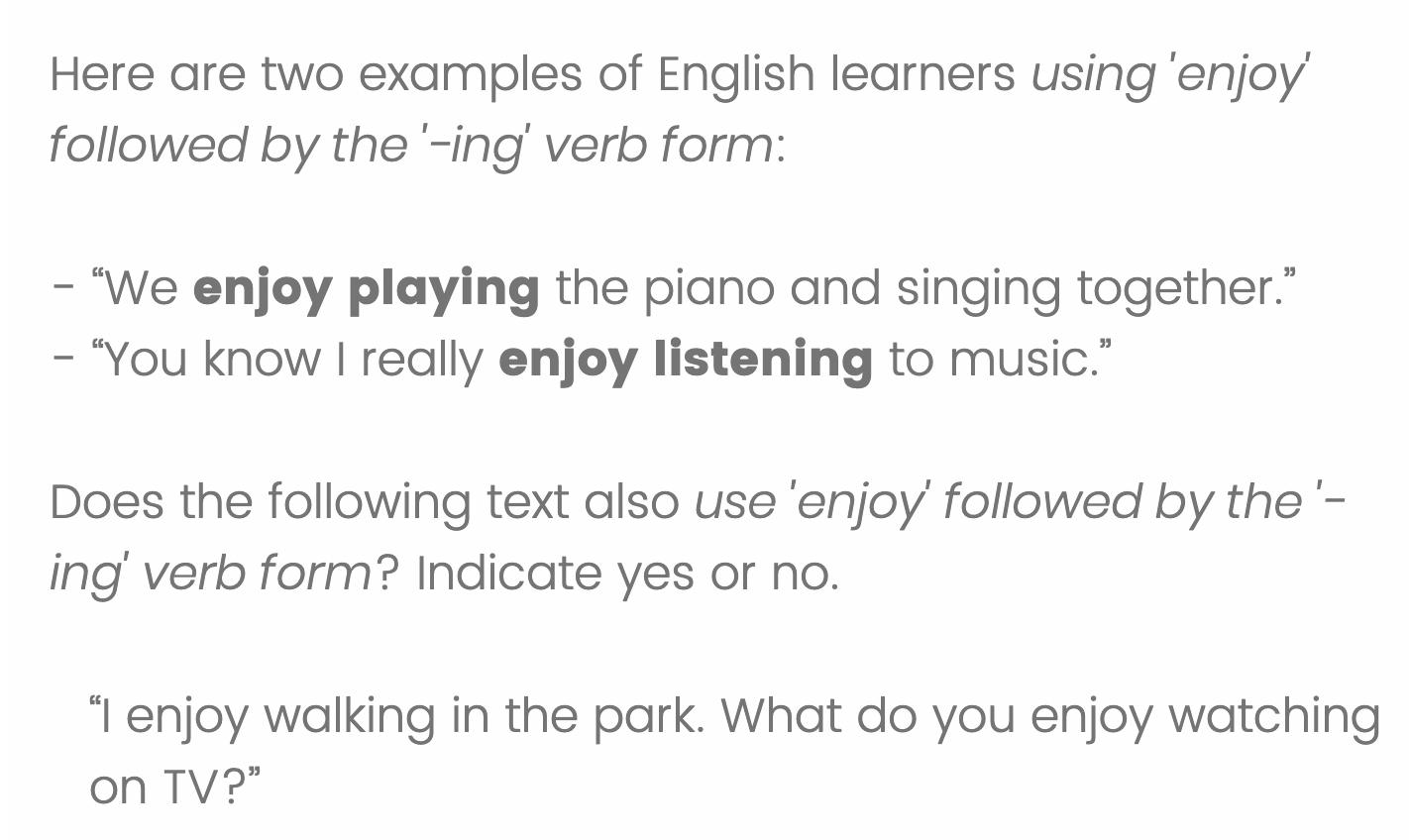}
    \caption{Example grammar evaluation task for Experiment 1.}
    \label{fig:human-e1-grammar}
  \end{figure}

\begin{figure}
    \centering
    \includegraphics[width=\linewidth]{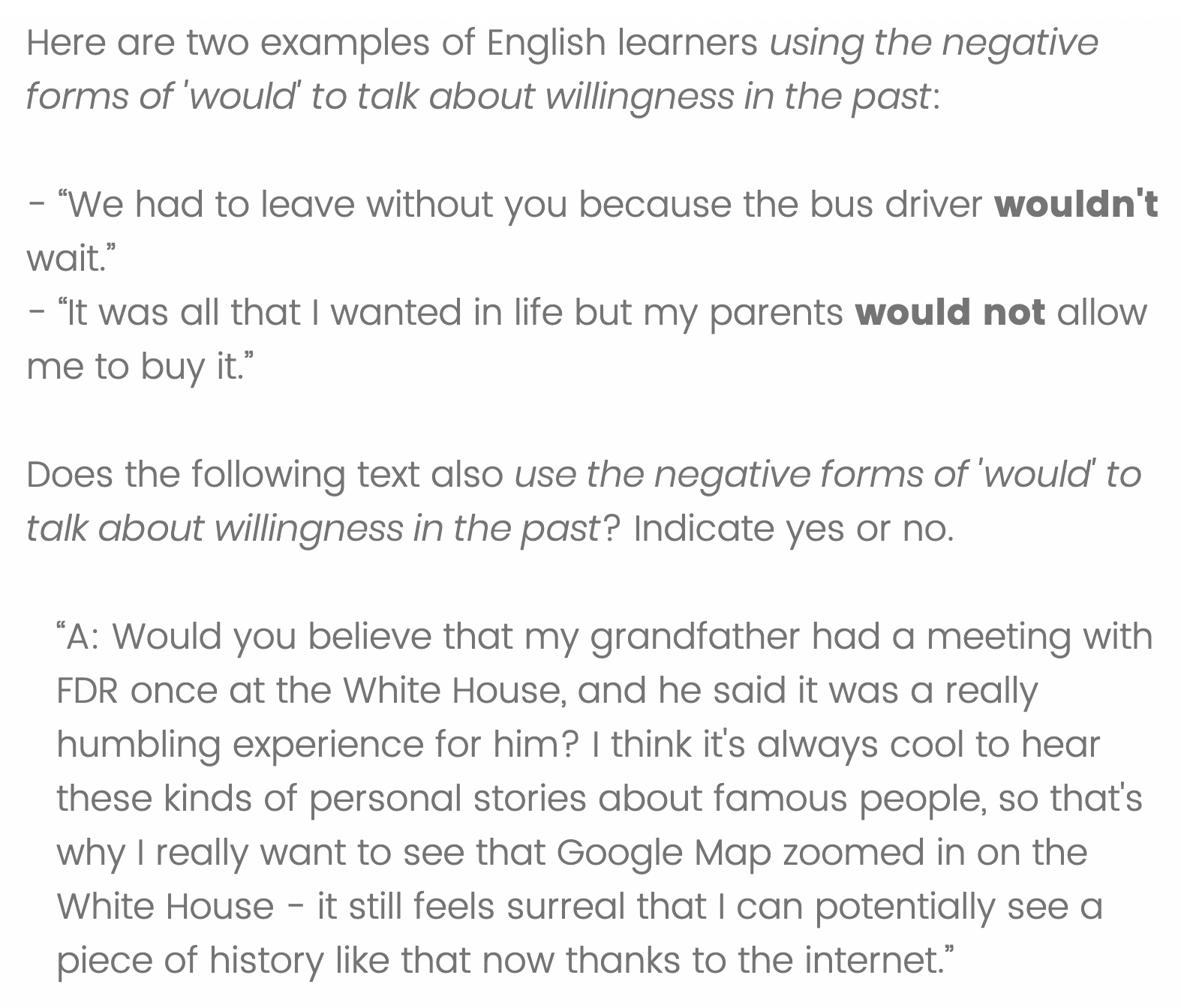}
    \caption{Example grammar evaluation task for Experiment 2. Raters are asked to evaluate one grammar target at a time; generations with multiple grammar targets are presented multiple times so raters can independently assess each target.}
    \label{fig:human-e2-grammar}
  \end{figure}

  We run two separate surveys: one for grammar evaluation, and one for quality evaluation. Both use the same set of sampled items from each experiment. For grammar evaluation, we provide a lightweight setup parallel to the LLM evaluation; however, where the LLM receives the entire EGP specification, the human raters instead see a minimal overview which names the target grammar structure and provides two example EGP learner sentences with the target grammar structure emphasized in bold (\Cref{fig:human-e1-grammar}). In Experiment 2, some generations have multiple grammar targets---in these cases we present the same generation separately to the rater for independent evaluation of each grammar target (\Cref{fig:human-e2-grammar}). We include four attention checks in the grammar survey, and retain only raters who pass at least three attention checks (11 participants total).

\begin{figure}
    \centering
    \includegraphics[width=\linewidth]{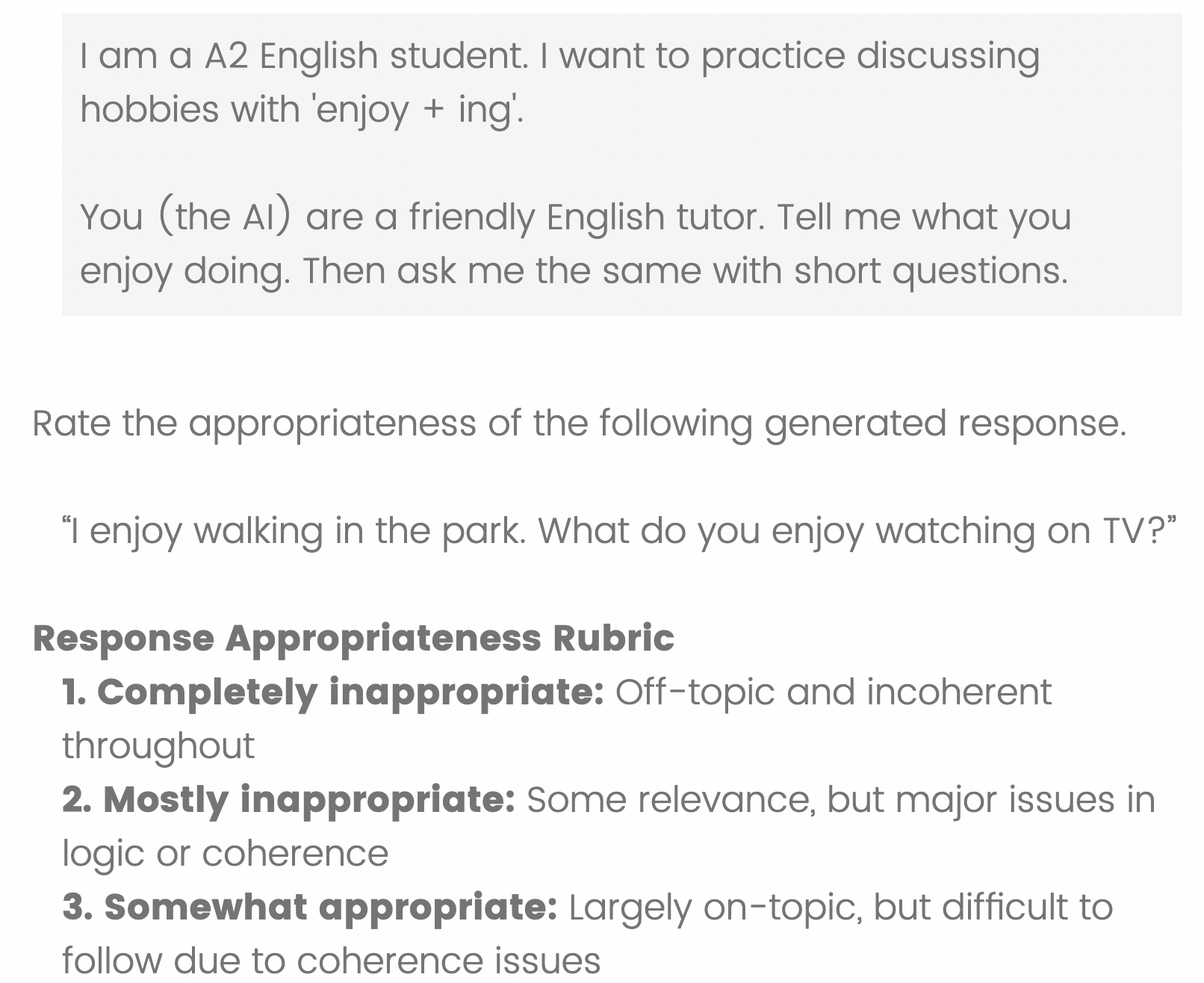}
    \caption{Example quality evaluation task for Experiment 1. Note that the quality response rubric here is cut off, but goes to level 5. We use the same quality rubric as \citet{ribeiro-flucht_framework_2025}.}
    \label{fig:human-e1-qual}
  \end{figure}

\begin{figure}
    \centering
    \includegraphics[width=\linewidth]{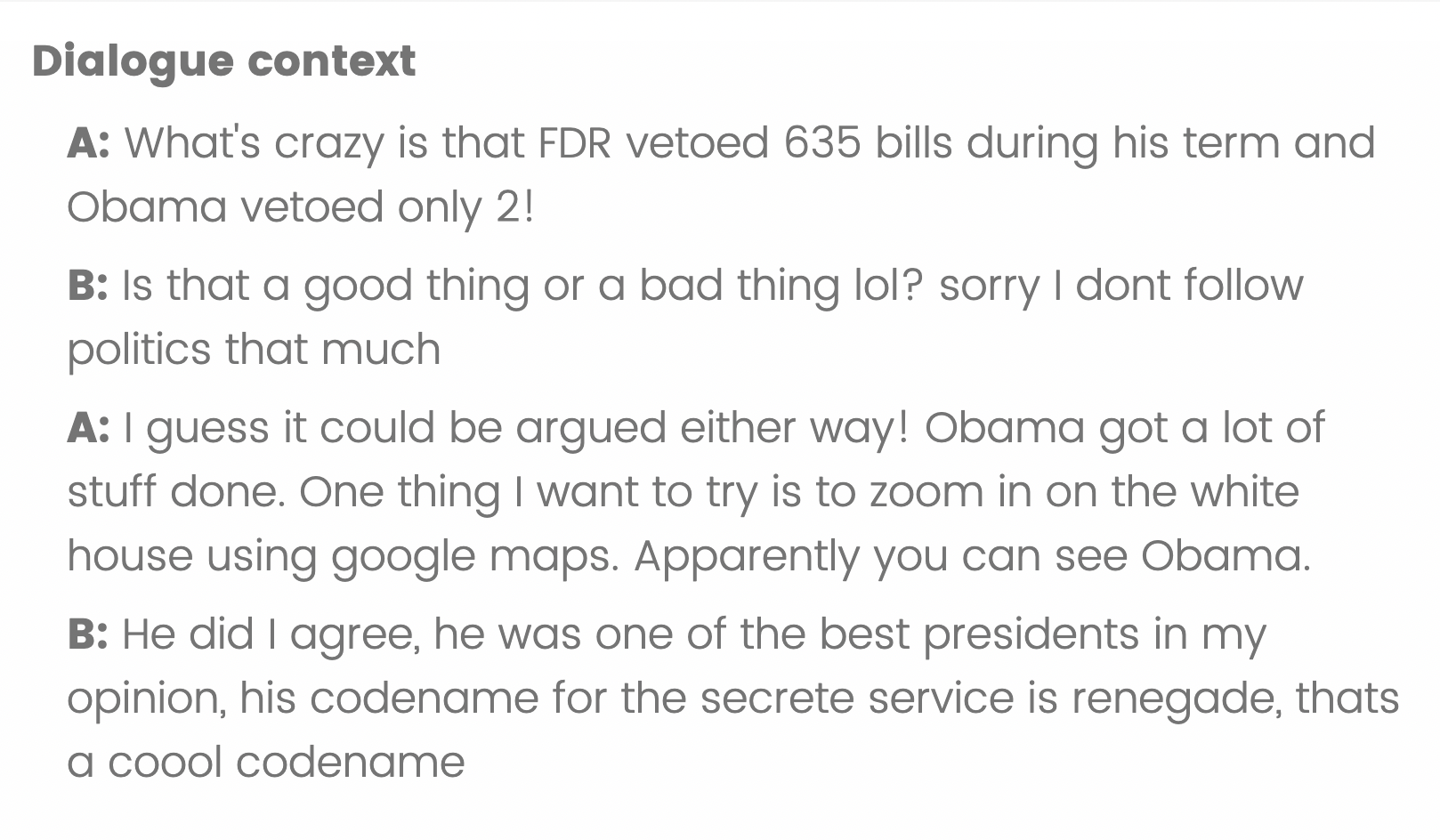}
    \includegraphics[width=\linewidth]{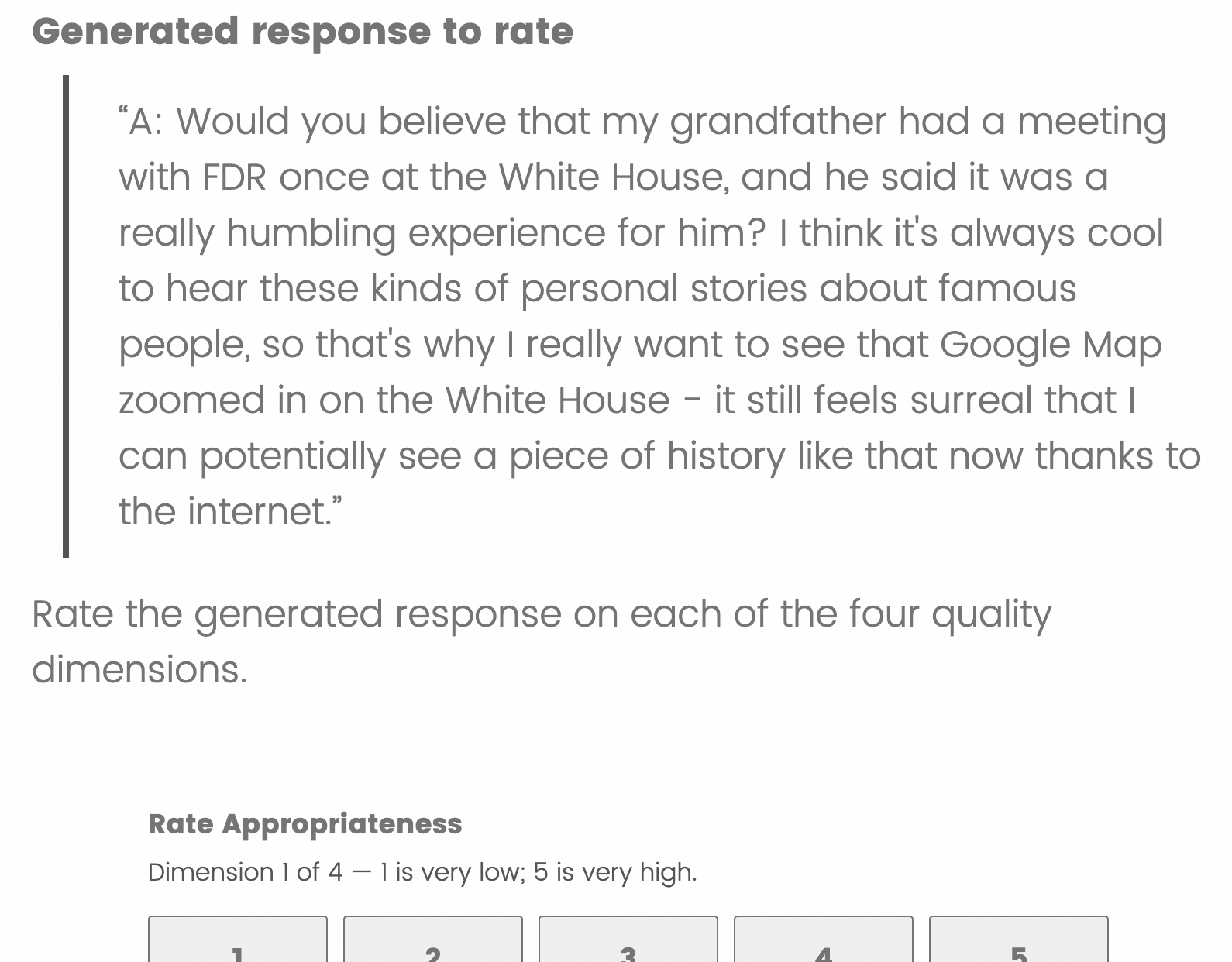}
    \caption{Example quality evaluation task for Experiment 2. Raters see the preceding dialogue context, and are prompted to independently rate the four quality dimensions used by \citet{glandorf_grammar_2025}---after rating Appropriateness (as shown here), the participant is next prompted to rate Content Richness and so on.}
    \label{fig:human-e2-qual}
  \end{figure}

  The quality survey uses 1--5 rating scales parallel to the two original experiments. Generations from Experiment 1 are assessed with a provided rubric (\Cref{fig:human-e1-qual}). As Experiment 2 uses natural dialogue data, we provide the context of preceding dialogue turns and ask participants to rate each generation along four quality dimensions (\Cref{fig:human-e2-qual}). Due to experimenter error, 5 raters completed a version of the quality survey with no attention checks. We retain their responses along with 8 raters who passed at least one out of two attention checks (13 participants total).

  \paragraph{Results.} Overall, we find the human ratings slightly lower but directionally consistent with our evaluation pipeline.
  In Experiment 1, we see that increasing the SMC particle budget improves both target grammar accuracy (Table \ref{tab:human-grammar-expt1}) and perceived quality (Table \ref{tab:human-quality-expt1}), and this effect is most drastic for the smallest model, Qwen 2 0.5B. For the longer generations in Experiment 2, increasing the particle budget similarly improves grammar accuracy in both single- and multiple-target conditions (Table \ref{tab:human-grammar-expt2}). For the quality assessment, however, human evaluators assign lower quality ratings on average compared to the automated pipeline, and we do not see the same quality improvement from increasing particle counts as in Experiment 1 (Table \ref{tab:human-quality-expt2}). This contrasts with the automated grammar (\Cref{tab:auto-grammar}) and quality (\Cref{tab:auto-quality}) ratings on the same sampled generations, which are generally higher but show the same general trend of improvement with increased particle count, with the exception of quality ratings in Experiment 2.

  \section{Example Generations}
  \label{app:example_generations}

  Please see Tables \ref{tab:expt1-example-generations}, \ref{tab:expt2-example-generations}, and \ref{tab:expt3-example-generations} for example generations from Experiments 1, 2, and 3 respectively.

  \section{Memory profiling}
  \label{app:memory_profile}

  To account for device memory needs, we profile memory usage from sample generations on both GPU (Table \ref{tab:gpu-memory-profile}) and CPU (Table \ref{tab:cpu-memory-profile}). For the latter, we use a quantized version of Llama 3.1 8B; as GenLM does not yet support quantization, we simulate memory usage through parallel generation.

  \begin{table*}[t] \small
    \centering
    \begin{tabular}{p{0.27\linewidth}p{0.14\linewidth}p{0.32\linewidth}ccc}
      \toprule \textbf{Can-do statement}                                                & \textbf{Modeling Setup}      & \textbf{Generated Text}                                                                                                    & \rotatebox{75}{GPT-5-nano} & \rotatebox{75}{Claude Haiku 4.5} & \rotatebox{75}{Gold}                          \\
      \midrule

      Can use a range of degree adverbs to modify common gradable adjectives. & Llama 3.1 8B, $\numParticles{=}16$, B, A2 & \textit{My house is big and white. It has a long, bright front door. The garden is very green and clean.}                  & \cmark                   & \xmark                           & \cmark                     \\
      Can use the affirmative form with an increasing range of verbs.                   & Llama 3.2 1B, $\numParticles{=}1$, B, A2  & \textit{Yesterday, I was at school.}                                                                                       & \cmark                   & \xmark                           & \cmark                     \\
      Can use a range of degree adverbs to modify common gradable adjectives.           & Llama 3.1 8B, $\numParticles{=}16$, B, A2 & \textit{I'm an English teacher. I teach students how to read, write, and speak English very well and clearly.}             & \xmark                   & \cmark                           & \xmark                     \\
      Can use \textit{enjoy} followed by the \textit{-ing} form.                        & Qwen 2 1.5B, $\numParticles{=}32$, B, A2   & \textit{Certainly! [\ldots] How about going to the beach for a day? It's a great way to relax and enjoy the sun. [\ldots]} & \xmark                   & \cmark                           & \xmark                     \\

      Can use \textit{enjoy} followed by the \textit{-ing} form.              & Llama 3.2 1B, $\numParticles{=}8$, B, any & \textit{[\ldots] I'll respond with the next part of the question in ``enjoying'' + ``ing''.}                               & \cmark                   & \xmark                           & \xmark                     \\
      Can use the affirmative form with an increasing range of verbs.                   & Llama 3.2 1B, $\numParticles{=}16$, T, A2 & \textit{I used to play with balls in a park.}                                                                              & \xmark                   & \cmark                           & \cmark                     \\
      \bottomrule
    \end{tabular}
    \caption{Representative annotator disagreements from Experiment~1
    generations. The setup column indicates the model, SMC particle count $\numParticles$, prompt
    format (B: basic, T: template), and vocabulary restriction (A2-only / any).}
    \label{tab:annotator-disagreements}
  \end{table*}

  \begin{table*}[ht]
    \centering
    \small
    \begin{tabular}{|@{}c@{}|}
      \hline
      \begin{minipage}{\dimexpr\textwidth-2\tabcolsep-2\arrayrulewidth}
        \vspace{8pt}
        \setlength{\parindent}{0pt}
        \setlength{\parskip}{0pt}
        \raggedright

        \textbf{WHAT IS THE GOAL OF THIS STUDY AND WHAT WILL I DO?}

        \vspace{\baselineskip}

        This evaluation study is conducted by researchers at ETH Z\"urich, Saarland University, and other partner universities.

        \vspace{\baselineskip}

        \textbf{[Grammar]} The goal is to evaluate whether short English texts contain specific grammatical patterns relevant to language learning. You were invited because you are a fluent English speaker.

        \vspace{\baselineskip}

        If you choose to participate, you will read brief example sentences written by second language learners of English, showing correct usage of a target grammatical pattern.

        You will then judge whether a second text generated by a language model contains the same grammatical pattern.

        Note that the English learner example sentences demonstrate the specific target grammar form correctly, but may be awkward or ungrammatical in other ways.

        \vspace{\baselineskip}

        Each judgment should take only a few seconds. The full study should take about 10--15 minutes. You will be compensated through Prolific.

        There are no risks associated with your participation.

        \vspace{\baselineskip}

        \textbf{[Quality]} The goal is to evaluate the quality of short English texts generated by language models.

        \vspace{\baselineskip}

        You will rate generated texts on 1--5 scales. The full study should take about 10--15 minutes. You will be compensated through Prolific.

        \vspace{\baselineskip}

        \textbf{USE OF YOUR DATA}

        \vspace{\baselineskip}

        \textbf{[Grammar]} We will ask for your Prolific ID and your yes/no judgments only. We do not collect personally identifying information.

        \textbf{[Quality]} We will ask for your Prolific ID and your quality ratings only.

        \vspace{\baselineskip}

        Data will be anonymised and may be published for research purposes.

        \vspace{\baselineskip}

        \textbf{IF YOU WISH TO WITHDRAW}

        \vspace{\baselineskip}

        You may leave the study at any time on Prolific. If you withdraw before completing the study, your data will be deleted and you will not receive payment.

        \vspace{8pt}
      \end{minipage}
      \\
      \hline
    \end{tabular}
    \caption{Informed consent page shown to participants. \textbf{[Grammar]} / \textbf{[Quality]} introduce the wording used in those respective surveys.}
    \label{tab:human-consent}
  \end{table*}

  \begin{table*}[h]
    \centering
    \small
    \begin{tabular}{|@{}c@{}|}
      \hline
      \begin{minipage}{\dimexpr\textwidth-2\tabcolsep-2\arrayrulewidth}
        \vspace{8pt}
        \setlength{\parindent}{0pt}
        \setlength{\parskip}{0pt}
        \raggedright

        \textbf{Instructions}

        \vspace{\baselineskip}

        \textbf{[Grammar]} You will see many short items like the practice items on the next pages.

        \vspace{\baselineskip}

        For each item:

        1. Read the two example sentences produced by learners of English as a second language.

        2. Read the generated text.

        3. Answer \textbf{Yes} if the generated text contains at least one instance of the target grammar pattern; otherwise answer \textbf{No}.

        You can press \textbf{Y} or \textbf{F} for Yes and \textbf{N} or \textbf{J} for No on your keyboard.

        \vspace{\baselineskip}

        Answer from your own judgment only. Do \textbf{not} use translation tools, dictionaries, search engines, or AI assistants (including ChatGPT). These items are designed to be answered quickly.

        \vspace{\baselineskip}

        Work at a steady pace and do not overthink individual items.

        \vspace{\baselineskip}

        \textbf{[Quality]} You will rate the quality of generated English texts on scales from 1 to 5.

        \vspace{\baselineskip}

        In the first section, give one appropriateness rating per text. In the second section, give four ratings per text: Appropriateness, Relevance, Content Richness, and Grammatical Correctness.

        \vspace{\baselineskip}

        Click a numbered button, or press a number key from \textbf{1} to \textbf{5}. You can also use \textbf{C, V, B, N, M} for ratings 1, 2, 3, 4, and 5. Each response is recorded immediately.

        \vspace{\baselineskip}

        Answer from your own judgment only. Do not use translation tools, search engines, or AI assistants.

        \vspace{8pt}
      \end{minipage}
      \\
      \hline
    \end{tabular}
    \caption{Task instructions page shown to participants. \textbf{[Grammar]} / \textbf{[Quality]} indicate the wording used in those respective surveys.}
    \label{tab:human-instructions}
  \end{table*}

  \begin{table*}[ht]
    \centering
    \small
    \begin{subtable}{0.47\textwidth}
      \centering

      \begin{tabular}{lrrr}
        \toprule
        & \textbf{N. particles} &  &  \\
        \textbf{Model} & 1 & 4 & 16 \\
        \midrule
        \multicolumn{4}{l}{\textbf{Unrestricted vocabulary}} \\
        Llama 3.1 8B & 95.5\% & 100.0\% & 95.5\% \\
        Qwen 2 0.5B & 34.8\% & 80.3\% & 98.5\% \\
        \midrule
        \multicolumn{4}{l}{\textbf{A2-only vocabulary}} \\
        Llama 3.1 8B & 81.8\% & 98.5\% & 90.9\% \\
        Qwen 2 0.5B & 39.4\% & 65.2\% & 97.0\% \\
        \bottomrule
      \end{tabular}
      \caption{Experiment~1 (6 items per model--particle condition). Columns are number of SMC particles.}
      \label{tab:human-grammar-expt1}
    \end{subtable}
    \hfill
    \begin{subtable}{0.47\textwidth}
      \centering
      \begin{tabular}{lrrr}
        \toprule
        & \textbf{N. targets} &  &  \\
        \textbf{N. particles} & 1 & 3 & 6 \\
        \midrule
        2 & 81.8\% & 48.5\% & 42.9\% \\
        8 & 78.8\% & 55.6\% & 51.0\% \\
        32 & 87.9\% & 66.7\% & 60.1\% \\
        \bottomrule
      \end{tabular}
      \caption{Experiment~2 (Llama 3.1 8B, 3 items per condition).}
      \label{tab:human-grammar-expt2}
    \end{subtable}
    \caption{Human-evaluated grammar target accuracy ($n=11$ participants). }

  \end{table*}

  \begin{table*}[ht]
    \centering
    \small
    \begin{subtable}{0.47\textwidth}
      \centering
      \begin{tabular}{lrrr}
        \toprule
        & \textbf{N. particles} &  &  \\
        \textbf{Model} & 1 & 4 & 16 \\
        \midrule
        \multicolumn{4}{l}{\textbf{Unrestricted vocabulary}} \\
        Llama 3.1 8B & 100.0\% & 100.0\% & 100.0\% \\
        Qwen 2 0.5B & 50.0\% & 83.3\% & 100.0\% \\
        \midrule
        \multicolumn{4}{l}{\textbf{A2-only vocabulary}} \\
        Llama 3.1 8B & 83.3\% & 100.0\% & 100.0\% \\
        Qwen 2 0.5B & 50.0\% & 83.3\% & 100.0\% \\
        \bottomrule
      \end{tabular}
      \caption{Experiment~1 (6 items per model--particle condition). Columns are number of SMC particles.}
      \label{tab:auto-grammar-expt1}
    \end{subtable}
    \hfill
    \begin{subtable}{0.47\textwidth}
      \centering
      \begin{tabular}{lrrr}
        \toprule
        & \textbf{N. targets} &  &  \\
        \textbf{N. particles} & 1 & 3 & 6 \\
        \midrule
        2 & 100.0\% & 55.6\% & 27.8\% \\
        8 & 100.0\% & 77.8\% & 61.1\% \\
        32 & 100.0\% & 77.8\% & 66.7\% \\
        \bottomrule
      \end{tabular}
      \caption{Experiment~2 (Llama 3.1 8B, 3 items per condition).}
      \label{tab:auto-grammar-expt2}
    \end{subtable}
    \caption{Automated grammar target accuracy on the human-evaluation sample.}
    \label{tab:auto-grammar}
  \end{table*}

  \begin{table*}[h]
    \centering
    \small
    \begin{subtable}{0.47\textwidth}
      \centering

      \begin{tabular}{lrrr}
        \toprule
        & \textbf{N. particles} &  &  \\
        \textbf{Model} & 1 & 4 & 16 \\
        \midrule
        \multicolumn{4}{l}{\textbf{Unrestricted vocabulary}} \\
        Llama 3.1 8B & 3.17 & 3.59 & 4.08 \\
        Qwen 2 0.5B & 2.33 & 2.77 & 3.59 \\
        \midrule
        \multicolumn{4}{l}{\textbf{A2-only vocabulary}} \\
        Llama 3.1 8B & 2.73 & 3.53 & 3.47 \\
        Qwen 2 0.5B & 2.05 & 2.08 & 2.64 \\
        \bottomrule
      \end{tabular}
      \caption{Experiment~1 (6 items per model--particle condition). Appropriateness only (1--5).}
      \label{tab:human-quality-expt1}
    \end{subtable}
    \hfill
    \begin{subtable}{0.47\textwidth}
      \centering
      \begin{tabular}{lrrrr}
        \toprule
        & \textbf{Metric} &  &  &  \\
        \textbf{N. particles} & App. & Rel. & CR & GC \\
        \midrule
        2 & 3.71 & 3.79 & 3.74 & 4.21 \\
        8 & 3.58 & 3.83 & 3.71 & 4.11 \\
        32 & 3.51 & 3.73 & 3.75 & 4.15 \\
        \bottomrule
      \end{tabular}
      \caption{Experiment~2 (Llama 3.1 8B, 8--9 items per particle condition). App.=appropriateness; Rel.=relevance; CR=content richness; GC=grammatical correctness.}
      \label{tab:human-quality-expt2}
    \end{subtable}
    \caption{Human-evaluated quality ratings ($n=13$ participants). Mean ratings on a 1--5 scale.}
    \label{tab:human-quality}
  \end{table*}

  \begin{table*}[h]
    \centering
    \small
    \begin{subtable}{0.47\textwidth}
      \centering
      \begin{tabular}{lrrr}
        \toprule
        & \textbf{N. particles} &  &  \\
        \textbf{Model} & 1 & 4 & 16 \\
        \midrule
        \multicolumn{4}{l}{\textbf{Unrestricted vocabulary}} \\
        Llama 3.1 8B & 5.00 & 4.67 & 5.00 \\
        Qwen 2 0.5B & 2.17 & 2.00 & 4.60 \\
        \midrule
        \multicolumn{4}{l}{\textbf{A2-only vocabulary}} \\
        Llama 3.1 8B & 3.67 & 4.67 & 5.00 \\
        Qwen 2 0.5B & 2.67 & 2.67 & 3.40 \\
        \bottomrule
      \end{tabular}
      \caption{Experiment~1 (6 items per model--particle condition). Appropriateness only (1--5).}
      \label{tab:auto-quality-expt1}
    \end{subtable}
    \hfill
    \begin{subtable}{0.47\textwidth}
      \centering
      \begin{tabular}{lrrrr}
        \toprule
        & \textbf{Metric} &  &  &  \\
        \textbf{N. particles} & App. & Rel. & CR & GC \\
        \midrule
        2 & 4.00 & 4.11 & 3.11 & 4.22 \\
        8 & 3.11 & 3.22 & 2.67 & 4.22 \\
        32 & 3.56 & 4.00 & 2.78 & 4.67 \\
        \bottomrule
      \end{tabular}
      \caption{Experiment~2 (Llama 3.1 8B, 9 items per particle condition). App.=appropriateness; Rel.=relevance; CR=content richness; GC=grammatical correctness.}
      \label{tab:auto-quality-expt2}
    \end{subtable}
    \caption{Automated quality ratings on the human-evaluation sample. Mean ratings on a 1--5 scale.}
    \label{tab:auto-quality}
  \end{table*}

  \begin{figure*}[ht!]
    \centering
    \includegraphics[width=0.95\linewidth]{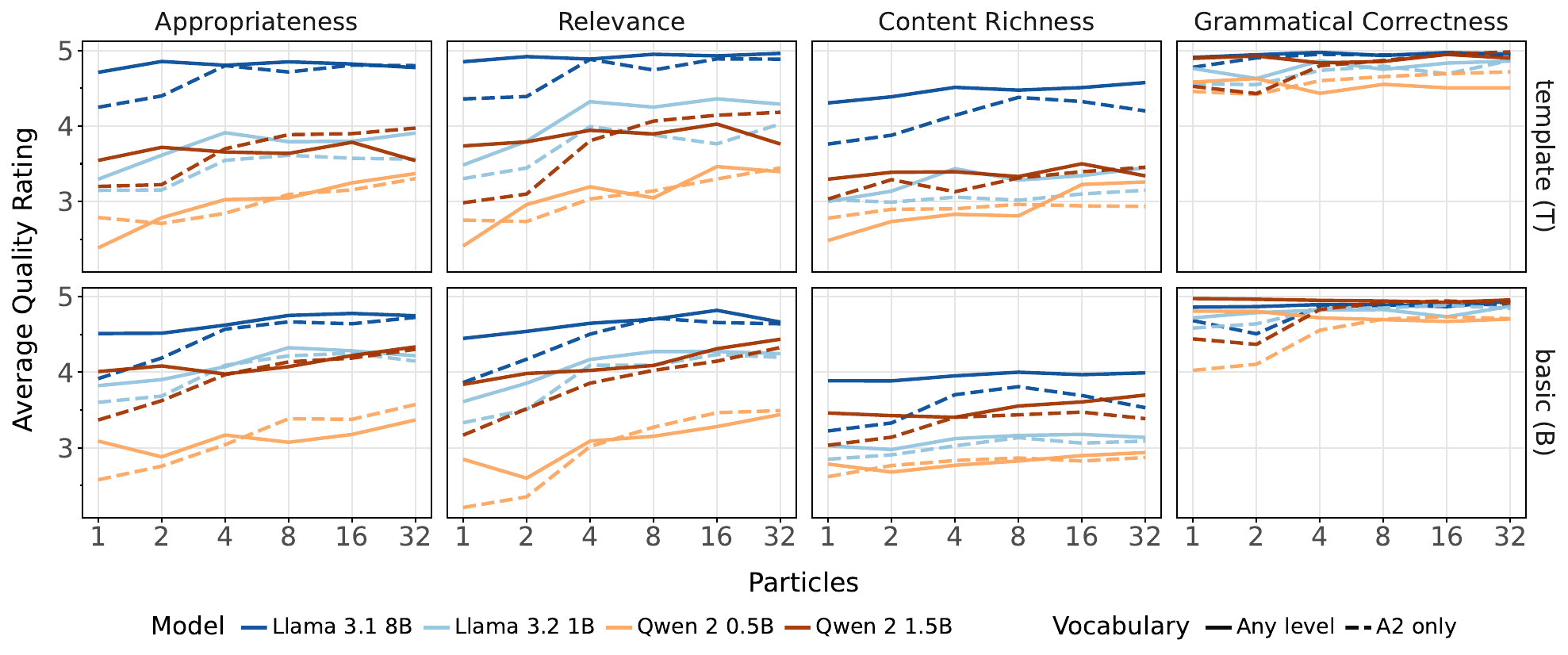}
    \caption{Experiment 1: Average quality metrics by number of SMC particles.}
    \label{fig:expt1_quality}
  \end{figure*}

  \begin{table*}[t]
    \centering
    \small
    \setlength{\tabcolsep}{3.5pt}
    \begin{tabular}{@{}lllrrrrrr@{}}
      \toprule
      \multicolumn{3}{c}{} &
      \multicolumn{6}{c}{Target-grammar coverage (\%)} \\
      \cmidrule(lr){4-9}
      Experiment & Model & Prompt &
      $\numParticles{=}1$ & $\numParticles{=}2$ & $\numParticles{=}4$ & $\numParticles{=}8$ & $\numParticles{=}16$ & $\numParticles{=}32$ \\
      \midrule
      1 & Llama 3.1 8B & B  & 32.0 & 46.4 & 60.8 & 72.8 & 60.8 & 64.0 \\
      &               & T     & 92.8 & 92.8 & 92.0 & 94.4 & 96.8 & 90.4 \\
      & Llama 3.2 1B & B     & 17.6 & 47.2 & 64.0 & 76.8 & 78.4 & 86.4 \\
      &               & T     & 56.8 & 80.0 & 89.6 & 92.0 & 91.2 & 98.4 \\
      & Qwen 2 0.5B  & B     & 12.8 & 28.8 & 31.2 & 56.8 & 64.8 & 84.0 \\
      &               & T     & 36.0 & 46.4 & 69.6 & 82.4 & 88.8 & 91.2 \\
      & Qwen 2 1.5B  & B     & 12.0 & 13.6 & 28.8 & 44.0 & 60.0 & 67.2 \\
      &               & T     & 27.2 & 46.4 & 74.4 & 85.6 & 94.4 & 95.2 \\
      \midrule
      2 & Llama 3.2 1B &   & 19.1 & 28.7 & 34.3 & 42.0 & 50.1 & 55.7 \\
      & Qwen 2 0.5B  &   &  4.1 &  7.1 & 12.7 & 19.6 & 28.3 & 37.1 \\
      & Qwen 2 1.5B  &   &  5.0 &  8.2 & 13.8 & 20.7 & 26.5 & 33.6 \\
      \bottomrule
    \end{tabular}
    \caption{Rejection-sampling results. Rejection failures without a selected string count as zero
    coverage, while runs with no terminated string use their highest-weight
    particle. Experiment~1 reports target grammar
    coverage for basic (B) and template (T) prompts across 125 task sets per
    configuration. Experiment~2 reports the percentage of 1,322 dialogue cases with at
    least one terminated completion that passed every \texttt{complete()} check.}
    \label{tab:rejection-sampling}
  \end{table*}

  \begin{table*}[t]
    \centering
    \small
    \setlength{\tabcolsep}{4pt}
    \begin{tabular}{@{}lllrrrrrr@{}}
      \toprule
      Experiment & Model & Peak memory (GiB) &
      $\numParticles{=}1$ & $\numParticles{=}2$ & $\numParticles{=}4$ & $\numParticles{=}8$ & $\numParticles{=}16$ & $\numParticles{=}32$ \\
      \midrule
      1 & Llama 3.1 8B & Host RSS & 5.64 & 5.59 & 5.35 & 5.21 & 5.28 & 5.50 \\
      &               & GPU VRAM & 15.84 & 16.95 & 19.66 & 26.23 & 39.44 & 39.08 \\
      & Qwen 2 0.5B  & Host RSS & 1.90 & 1.81 & 1.85 & 1.99 & 1.99 & 2.34 \\
      &               & GPU VRAM & 1.60 & 2.00 & 2.78 & 15.45 & 8.91 & 35.54 \\
      \midrule
      2 & Llama 3.1 8B & Host RSS & 5.48 & 5.53 & 4.88 & 5.59 & 5.10 & 5.67 \\
      &               & GPU VRAM & 16.55 & 19.05 & 24.23 & 38.91 & 39.36 & 39.16 \\
      & Qwen 2 0.5B  & Host RSS & 1.87 & 1.98 & 1.97 & 2.14 & 2.23 & 2.16 \\
      &               & GPU VRAM & 1.74 & 4.60 & 5.23 & 15.25 & 9.79 & 27.08 \\
      \bottomrule
    \end{tabular}
    \caption{Peak memory (Resident Set Size, i.e., physical RAM usage, and GPU VRAM) for \textbf{GPU} SMC runs with a 100-token generation budget for our largest and smallest models on an A100 with 40\,GiB.
    Experiment~1 profiles one generation task, while Experiment~2 reports peak memory usage over the
    first three tasks, %
    so these are stochastic assessments rather than worst-case bounds.
    }\label{tab:gpu-memory-profile}
  \end{table*}

  \begin{table*}[t]
    \centering
    \small
    \setlength{\tabcolsep}{4pt}
    \begin{tabular}{@{}llllrrrrrr@{}}
      \toprule
      Experiment & Model & Tested & Quantization &
      $\numParticles{=}1$ & $\numParticles{=}2$ & $\numParticles{=}4$ & $\numParticles{=}8$ & $\numParticles{=}16$ & $\numParticles{=}32$ \\
      \midrule
      1 & Qwen 2 0.5B  &  SMC           & None      & 2.31 & 2.07 & 2.72 & 2.95 & 3.36 & 4.96 \\
      2 & Qwen 2 0.5B  &  SMC           & None      & 2.34 & 2.41 & 2.96 & 4.63 & 3.71 & 7.49 \\
      1 & Llama 3.1 8B & standard baseline  & Q4\_K\_M  & 7.87 & 7.92 & 7.98 & 8.18 & 8.46  & 9.00 \\ %
      \bottomrule
    \end{tabular}
    \caption{Peak memory (process-tree Resident Set Size) in GiB for \textbf{CPU}-only runs with a 100-token
    budget, comparing the smallest Qwen model and quantized Llama. $\numParticles$ is the particle count for Qwen SMC. As GenLM does not yet support quantized models, the Llama model instead runs the standard language model baseline with
    Q4\_K\_M quantization and $\numParticles$ parallel generations for four CPU threads. }\label{tab:cpu-memory-profile}
  \end{table*}

  \begin{table*}[ht]
    \centering
    \small
    \begin{tabular}{@{}p{0.96\textwidth}@{}}
      \toprule
      \textbf{SYSTEM:} Only output A's response.\\
      \addlinespace[0.6ex]
      \textbf{USER:} Given the dialog, write a possible next turn of A that includes all of these grammatical items:
      \begin{itemize}
        \setlength{\itemsep}{0.15em}
        \setlength{\parsep}{0pt}
        \setlength{\topsep}{0.2em}
        \item negation---FORM: QUESTIONS: Can use negative question forms in main clauses and question tags. $\blacktriangleright$ questions (CEFR B1)
        \item superlatives---FORM/USE: COMPLEX NOUN PHRASES: Can form a limited range of complex noun phrases with a superlative adjective + prepositional phrase, to talk about something unique. $\blacktriangleright$ noun phrases $\blacktriangleright$ clauses: comparison (CEFR A2)
        \item would---USE: WISHES AND PREFERENCES: Can use \emph{would} with a wide range of verbs to talk about wishes and preferences. (CEFR A2)
      \end{itemize}\\
      \addlinespace[0.4ex]
      \textit{Dialog:}\newline
      A: Hello have you seen the shape of water?\newline
      B: No, have you?\newline
      A: Yes, it's a fantasy/drama film, it's really good\newline
      B: Yeah? Who directed it again?\\
      \bottomrule
    \end{tabular}
    \caption{Example prompt for Experiment~2 (three simultaneous grammar constraints).}
    \label{tab:expt2_prompt}
  \end{table*}

  \begin{table*}[t]
    \centering
    \small
    \setlength{\tabcolsep}{4pt}
    \begin{tabular}{@{}p{0.17\textwidth} p{0.38\textwidth} p{0.38\textwidth}@{}}
      \toprule
      \textbf{Grammar target} & \textbf{Unrestricted vocabulary} & \textbf{A2 vocabulary} \\
      \midrule
      Degree adverb + adjective &
      1.\ I love ice cream. I think it's extremely cold and very sweet. What about burgers? Do you think burgers are quite big?\newline
      2.\ I have two very nice sisters. My mum is very kind. My dad is very successful. Do you have any siblings? Are they really interesting? &
      1.\ I have one sister, she is very kind. My mom is really helpful. My dad is so funny. And I have one dog, he is quite old.\newline
      2.\ My garden is very big. It's full of flowers and trees. My kitchen is quite old, but it's very clean. \newline Is your house old too? \\
      \midrule
      Past simple (affirmative) &
      1.\ I visited my family last month. We ate pizza together. What about last month? Did you eat at home or at a restaurant?\newline
      2.\ Last year was great. I studied English hard. I read six books. I went to the beach three times. I met my penfriend in Italy. &
      1.\ I went to the supermarket. I bought some eggs. I made breakfast. I ate toast. I drank tea. \newline  What did you do yesterday morning?\newline
      2.\ I am happy to talk to you about last week. \newline I visited my parents on Sunday. I bought some flowers for my mother. They were very beautiful. \\
      \midrule
      Present simple + frequency adverb &
      1.\ I usually wake up at 6. I eat breakfast sometimes, but not every day. Do you have breakfast in the morning?\newline
      2.\ People in my country often eat vegetables 5 times a week. How about your country? &
      1.\ I usually go for a walk on Saturday morning. I always watch a movie on Sunday night. Sometimes I play with my dog.\newline
      2.\ In my country, people often eat breakfast at home and then lunch at work. Sometimes they order pizza for dinner.\newline  Where are you from? \\
      \midrule
      Empty subject \emph{it} &
      1.\ I am in London, UK.\newline It is 25 degrees Celsius here. How is the weather there?\newline (Waiting for your answer...)\newline
      2.\ I was in my office yesterday. \newline Here it was quiet and we are working on English lessons. \newline Yesterday was a busy day in office. How was it at home? &
      1.\ I'm in a small town in the mountains. It is cold here. How is the weather where you are?\newline
      2.\ I was on the beach. Here it was windy yesterday. How was it there? \\
      \midrule
      \emph{enjoy} + \emph{-ing} &
      1.\ I enjoy swimming in the ocean. Sometimes, I enjoy watching the sunset. Do you enjoy hiking or do you like lying on the beach?\newline
      2.\ I don't enjoy reading books at night before bed, it makes me sleepy. What kind of exercise do you enjoy doing in the morning? &
      1.\ When I am on holiday, I enjoy swimming in the sea. How about you? \newline Do you enjoy playing games on holiday, like surfing or tennis?\newline
      2.\ I enjoy swimming. I like it because it's good exercise and fun. What kind of sports do you enjoy, like cycling, dancing, or something else? \\
      \bottomrule
    \end{tabular}
    \caption{Experiment~1: Example tutor turns under grammar potentials alone (middle) and with an additional A2 vocabulary potential (right). Each cell shows two generations from distinct communicative scenarios for the same target constraint (16 SMC particles, template prompt).}
    \label{tab:expt1-example-generations}
  \end{table*}

  \begin{table*}[t]
    \centering
    \small
    \setlength{\tabcolsep}{4pt}
    \begin{tabular}{@{}c c p{0.30\textwidth} p{0.52\textwidth}@{}}
      \toprule
      \textbf{\#} & & \textbf{Grammar targets} & \textbf{Generation} \\
      \midrule
      \multirow{2}{*}{1} & A & NEGATION: Can form negative statements of main verbs in the present continuous and present perfect with 'be' and 'have' + 'not/n't'. & A: I'm not surprised about Carol Burnett, many people don't have a clue about that. What I can't understand is why some chefs aren't working on making affordable meals for people. \\
      & B & CLAUSES: Can form conditional subordinate clauses with 'if' + past simple, and would in the main clause, to talk about an imagined situation, often in the context of advice or opinion-giving. & A:  If that were the case, I think sculpture would qualify as an Olympic sport, but I'm not sure if it would have been included if chess were removed from the Olympic programme long ago. \\
      \midrule
      \multirow{2}{*}{3} & A & NEGATION: Can form negative statements of main verbs in the present simple with 'don't' + main verb.; ADJECTIVES: Can form an increasing range of complex noun phrases with a superlative adjective + prepositional phrase, to talk about something unique.; MODALITY: Can use 'would' to talk about imagined situations in the past. & I don't know if it's true, but I read that those cables have a lower transmission loss than any other material, that's probably why they're the best choice for that job, I'd imagine. \\
      & B & NEGATION: Can form negative statements of main verbs in the past simple with 'didn't' + main verb.; ADJECTIVES: Can use the irregular superlative adjective 'best' in the phrase 'my best friend'.; MODALITY: Can use 'would' to talk about the future in the past. & A:  Yeah, he was my best friend, but I never thought he would get caught, didn't he? I mean, he always said he'd retire a millionaire by the time he was 30, but I guess that didn't happen. \\
      \midrule
      \multirow{2}{*}{6} & A & NEGATION: Can use uncontracted 'not' for emphasis or in formal contexts.; NEGATION: Can use negative form 'can't'.; ADJECTIVES: Can form an increasing range of complex noun phrases with a superlative adjective + prepositional phrase, to talk about something unique.; ADJECTIVES: Can form a limited range of complex noun phrases with a superlative adjective + prepositional phrase, to talk about something unique.; MODALITY: Can use 'would like' to talk about wishes and preferences.; MODALITY: Can use question forms. & ``I would love to read that book, can't you just imagine sipping whiskey in a cup from the most romantic tea party in the south, that's not something I can see happening anytime soon! \\
      & B & NEGATION: Can form negative statements of main verbs in the present continuous and present perfect with 'be' and 'have' + 'not/n't'.; NEGATION: Can use uncontracted 'not' for emphasis or in formal contexts.; ADJECTIVES: Can form a limited range of complex noun phrases with a superlative adjective + prepositional phrase, to talk about something unique.; ADJECTIVES: Can form an increasing range of complex noun phrases with a superlative adjective + prepositional phrase, to talk about something unique.; MODALITY: Can use 'would' to talk about the future in the past.; MODALITY: Can use the affirmative form. & I would argue that 'Granny Style' isn't the best choice of shooting technique even if it's preferred by physicists,  I'm not sure it's the most effective method for pros in the league. \\
      \bottomrule
    \end{tabular}
    \caption{Experiment~2: Example dialogue continuations for one, three, and six simultaneous EGP constraints. Each block shows two distinct target constraint combinations (rows~A and~B) from the Task~1 benchmark (16 SMC particles).}
    \label{tab:expt2-example-generations}
  \end{table*}

  \begin{table*}[t]
    \centering
    \small
    \setlength{\tabcolsep}{4pt}
    \begin{tabular}{@{}c p{0.30\textwidth} p{0.52\textwidth}@{}}
      \toprule
      \textbf{CEFR level} & \textbf{EGP target} & \textbf{Generation} \\
      \midrule
      \multirow{2}{*}{A1} & CLAUSES: Can combine two main declarative clauses using co-ordinating conjunctions ('and', 'but', 'or'). & It's a heartwarming movie and very emotionally moving, especially in the way it portrays the relationships between the characters. \\
      & NOUNS: Can form noun phrases by pre-modifying a limited range of nouns with another noun. & A: That's interesting, but can you tell me more about the portrayal of British soldiers on warships like the HMS  Queen Elizabeth? \\
      \midrule
      \multirow{2}{*}{A2} & PRESENT: Can use the present simple after 'if' to talk about real and imagined situations. & A: I'm here. Don't worry, we can take our time. If we didn't have to meet, we probably would stay at home and watch another movie. \\
      & PRONOUNS: Can use 'this' as a pronoun to refer to something that is happening now. & A:  I completely agree, THIS is what makes the movie so special - the awesome connection between the actors and their characters. \\
      \midrule
      \multirow{2}{*}{B1} & PRONOUNS: Can use determiners + premodifiers + 'ones'. & Actually, I'm not really in the mood for anything right now, but ones that have action and adventure are usually my favorite ones. \\
      & MODALITY: Can use 'it' + 'be' + 'important' + ('that') clause. & B: Of course, lots of action and imagination.\newline A: Exactly, it is often very important that a movie has a well-structured narrative. \\
      \midrule
      \multirow{2}{*}{B2} & CLAUSES: Can combine two main clauses of the same type using 'neither' in front or mid position, followed by 'nor' to introduce the second clause, for emphasis. & Neither the Nighthowlers' appearance nor the timing of their debut in the movie is what initially caught my attention, I must say. \\
      & ADVERBS: Can use a wide range of adverbs and adverb phrases to refer to the timing of an event. & Slowly, I consider the possibility, but then I immediately decide it would be an excellent opportunity to shine on the big screen. \\
      \midrule
      \multirow{2}{*}{C1} & MODALITY: Can use 'be set to' to talk about strong assertion about the future, especially in institutional contexts. & A: Well, I guess that's what robots like me are always set to learn, new things and improve our film taste - Iron Man and Robocop. \\
      & FOCUS: Can use 'Wh-'cleft clauses as titles or subtitles, to point to something that follows, for focus. & What's often overlooked in this movie is the brilliant character development of Gru, which sets the tone for the rest of the film. \\
      \midrule
      \multirow{2}{*}{C2} & ADVERBS: Can use 'hardly' in front position followed by an inverted subject and verb, to give focus. & Hardly had I seen the movie where a robbery occurred from Chicago to Paris when I couldn't help but think of its implausible plot. \\
      & CONJUNCTIONS: Can use 'in that' as a subordinating conjunction, to give greater in-depth explanation, often in formal contexts. & A: apparently it's quite effective in that it uses satire to highlight social issues, which I think is a great way to address them. \\
      \bottomrule
    \end{tabular}
    \caption{Experiment~3: Example dialogue continuations for single EGP constraints across CEFR levels (two examples per level), using self-steered and repaired potentials (16 SMC particles).}
    \label{tab:expt3-example-generations}
  \end{table*}

\end{document}